\documentclass{article}

\usepackage{microtype}
\usepackage[final]{graphicx}
\usepackage{subfigure}
\usepackage{booktabs} %
\usepackage{placeins} %
\usepackage{multirow}
\usepackage{hyperref}

\usepackage[preprint]{icml2025}

\usepackage{amsmath}
\usepackage{amssymb}
\usepackage{mathtools}
\usepackage{amsthm}
\usepackage{caption}

\usepackage[capitalize,noabbrev]{cleveref}

\theoremstyle{plain}

\theoremstyle{definition}

\theoremstyle{remark}

\usepackage[disable,textsize=tiny]{todonotes}

\makeatletter
\renewcommand{\icmlaffiliation}[2]{}
\renewcommand{\@pa}[1]{}
\renewcommand{\printAffiliationsAndNotice}[1]{%
  \global\icml@noticeprintedtrue%
}
\makeatother

\icmltitlerunning{Light4D: Training-Free Extreme Viewpoint 4D Video Relighting}

\begin{document}

\twocolumn[
\icmltitle{Light4D: Training-Free Extreme Viewpoint 4D Video Relighting}

\begin{icmlauthorlist}
    \icmlauthor{Zhenghuang Wu$^{*}$}{}
    \icmlauthor{Kang Chen$^{*}$}{}
    \icmlauthor{Zeyu Zhang$^{*\dag}$}{}
    \icmlauthor{Hao Tang$^{\ddag}$}{}
\end{icmlauthorlist}

  \vspace{0.03in}
  \centerline{School of Computer Science, Peking University}
  \vspace{0.02in}
  \centerline{\footnotesize $^*$Equal contribution. $^\dag$Project lead. $^\ddag$Corresponding authors: bjdxtanghao@gmail.com.}

\makeatletter
\renewcommand{\@M}{
  \vspace*{1em} %
  \centering
  \includegraphics[width=\linewidth]{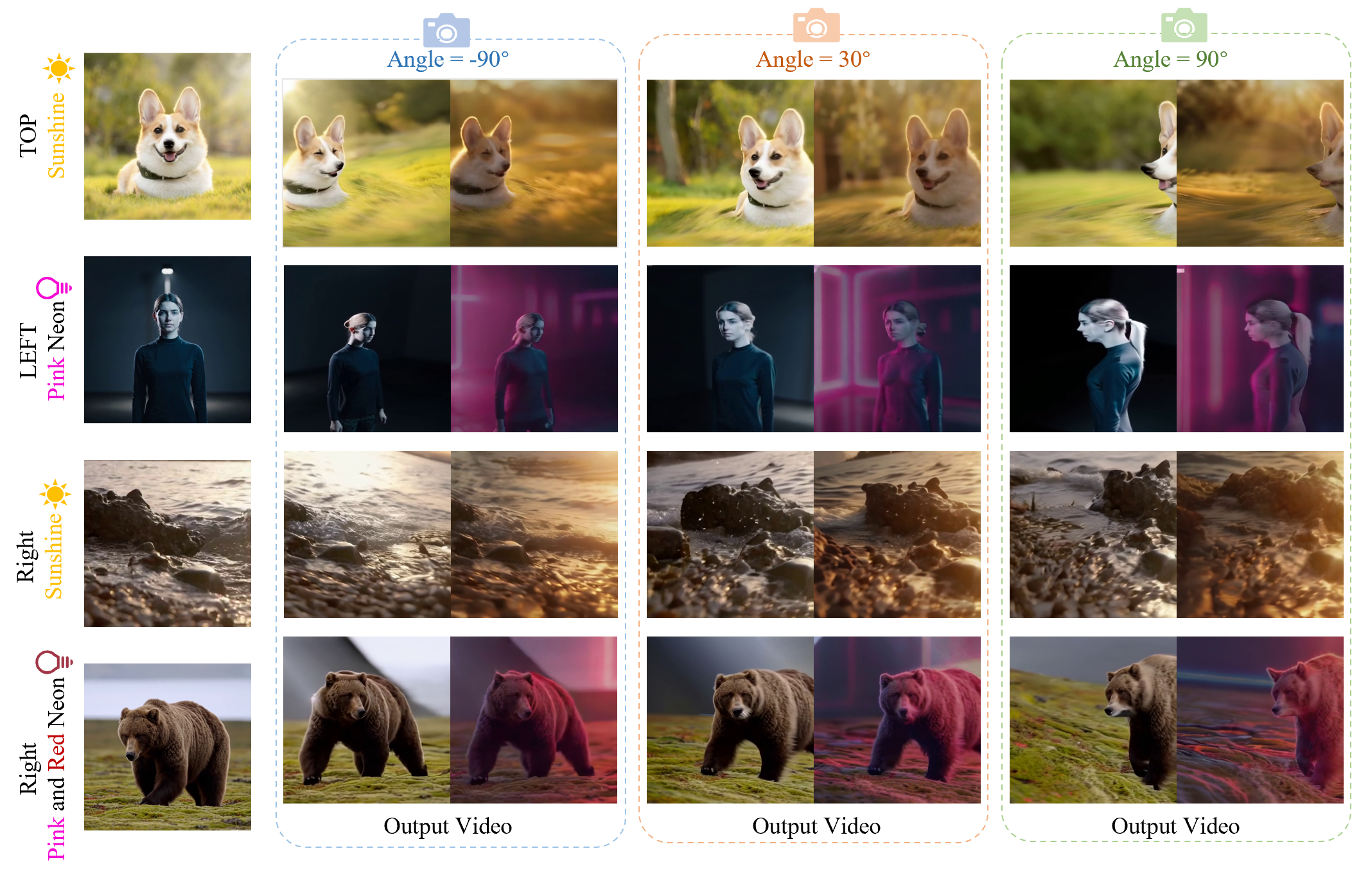}
  \captionof{figure}{\textbf{Visual results of Light4D for training-free 4D video relighting.} Our framework robustly handles extreme viewpoint changes and diverse lighting conditions while maintaining strict geometric-illumination consistency.}
  \label{fig:teaser}
  \bigskip
}
\makeatother

\icmlkeywords{Machine Learning, ICML}

\vskip 0.3in
]

\begin{abstract}
Recent advances in diffusion-based generative models have established a new paradigm for image and video relighting. However, extending these capabilities to 4D relighting remains challenging, due primarily to the scarcity of paired 4D relighting training data and the difficulty of maintaining temporal consistency across extreme viewpoints. In this work, we propose \textbf{\textit{Light4D}}, a novel training-free framework designed to synthesize consistent 4D videos under target illumination, even under extreme viewpoint changes. First, we introduce Disentangled Flow Guidance, a time-aware strategy that effectively injects lighting control into the latent space while preserving geometric integrity. Second, to reinforce temporal consistency, we develop Temporal Consistent Attention within the IC-Light architecture and further incorporate deterministic regularization to eliminate appearance flickering. Extensive experiments demonstrate that our method achieves competitive performance in temporal consistency and lighting fidelity, robustly handling camera rotations from $-90^{\circ}$ to $90^{\circ}$.
Code: \url{https://github.com/AIGeeksGroup/Light4D}.
Website: \url{https://aigeeksgroup.github.io/Light4D}. 
\end{abstract}

\section{Introduction}
The synthesis of dynamic 4D content is fundamental for next-generation immersive applications, including cinematic virtual production~\citep{bahmani20244d}, AR/VR~\citep{pang2025disco4d}, and interactive simulations~\citep{wen2025dynamicverse}. Since 4D geometric synthesis alone is insufficient for photorealism, achieving simultaneous control over both camera trajectory and illumination becomes a pivotal requirement for high-fidelity generation.

However, current research focuses predominantly on either video relighting or 4D geometric generation, leaving their intersection largely unexplored. Existing video relighting approaches~\citep{zhang2025scaling, lin2025illumicraft, bharadwaj2025genlit} operate primarily in the 2D domain. Although these methods improve the temporal consistency of videos with limited motion, they fundamentally struggle to maintain spatiotemporal consistency under complex camera trajectories. In contrast, camera-controlled 4D generation methods~\citep{chen20254dnex, mi2025one4d, zhou2025holotime} excel in synthesizing coherent dynamic geometry but typically overlook controllable illumination. In these frameworks, lighting effects are embedded in the texture~\citep{zhang2021nerfactor}, making the illumination uneditable and preventing adaptation to novel environments.

To address these limitations, recent research has explored 4D relighting via end-to-end supervised training. Frameworks such as Light-X~\citep{liu2025light} aim to learn joint camera and illumination control directly. However, these methods face two critical bottlenecks. First, they are fundamentally constrained by the severe data scarcity. Training these models necessitates massive paired datasets featuring both multi-view and multi-illumination consistency, which are prohibitively expensive to acquire in real-world environments. Second, these approaches often struggle to generalize to extreme viewpoints. Due to their reliance on limited synthetic data, the generated illumination tends to appear rigid and flat, resembling 2D texture mapping rather than volumetric light transport~\citep{chaturvedi2025synthlight}, which severely compromises photorealism.

In this work, we propose \textbf{Light4D}, a training-free framework tailored for 4D video relighting under extreme viewpoint changes. As shown in \cref{fig:teaser}, our method achieves high-fidelity results under extreme viewpoint changes. Unlike supervised methods that rely on expensive retraining, our method leverages the generative power of pre-trained expert models, synergizing the geometric prior of EX-4D~\citep{hu2025ex} with the illumination prior of IC-Light~\citep{zhang2025scaling}. A critical challenge in this integration arises from the inherent conflict between geometric reconstruction and illumination synthesis within the generative manifold. To resolve this, we introduce Disentangled Flow Guidance (DFG), a time-aware strategy that harmonizes lighting injection with geometric preservation. Specifically, this mechanism establishes a robust geometric foundation during the initial denoising phase, subsequently utilizing a lighting-fused latent state as a rectified target to steer the simultaneous synthesis of geometry and illumination. Furthermore, to ensure cross-frame coherence, we develop Temporal Consistent Attention (TCA) within the IC-Light architecture and employ deterministic regularization to suppress stochastic fluctuations and ensure temporal consistency.

Our contributions are summarized as follows:
\begin{itemize}
    \item We propose Light4D, the first training-free framework capable of achieving joint control over extreme camera trajectories ($-90^{\circ} \sim 90^{\circ}$) and illumination. By leveraging pre-trained generative priors, our approach eliminates the need for large-scale paired datasets.
    \item We introduce a Disentangled Flow Guidance strategy and Temporal Consistent Attention (TCA). These designs jointly resolve the inherent conflict between geometric reconstruction and illumination synthesis, ensuring strict temporal coherence and preventing structural degradation.
    \item Extensive experiments demonstrate that our method achieves competitive performance in terms of temporal consistency and lighting fidelity compared to existing video relighting baselines, particularly under extreme viewpoint changes.
\end{itemize}

\begin{figure*}[!ht]
\centering
\includegraphics[width=\textwidth]{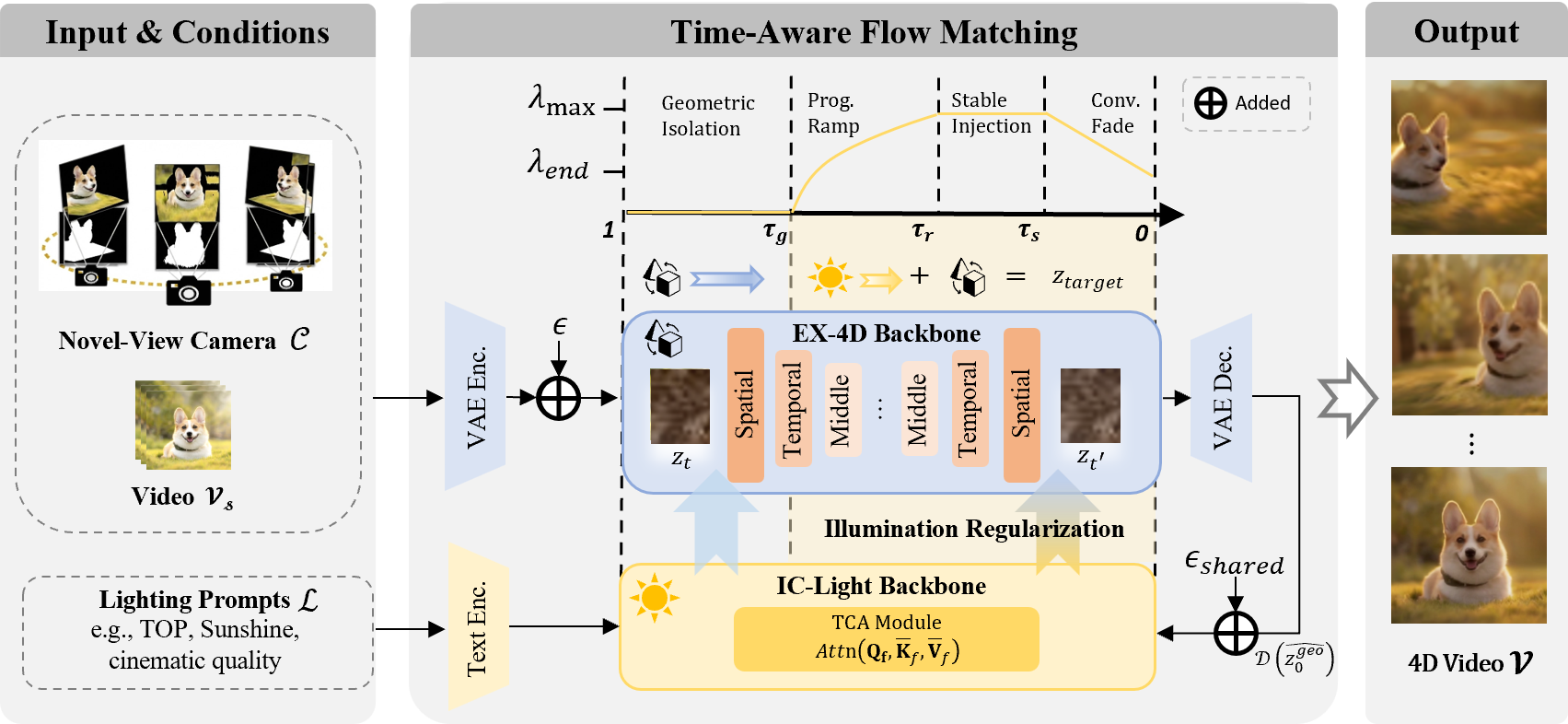}
\caption{\textbf{Overview of the Light4D framework.} Our training-free approach employs a time-aware paradigm in a latent flow-matching process. Using a multi-phase adaptive schedule $\lambda(t)$, we prioritize 3D geometric completion via the EX-4D backbone before injecting illumination cues through IC-Light.}
\label{fig:qualitative}
\vskip -0.2in
\end{figure*}

\section{Related Work}
\textbf{Learning-based Illumination Editing.} Early learning-based relighting methods typically rely on paired supervision, where encoder-decoder CNNs map an input image to a relit output. Extending relighting from images to videos often introduces temporal flicker, motivating explicit cross-frame consistency constraints \cite{zhang2021neural_video_portrait_relighting,chandran2022temporally}. In recent years, diffusion-based formulations have become a dominant direction \cite{ponglertnapakorn2023difareli,chaturvedi2025synthlight,wang2025comprehensive_relighting,fang2025relightvid,magar2025lightlab}. Their performance further benefits from scaling choices such as larger training corpora, stronger backbones, and physically grounded objectives and constraints \cite{kocsis2024lightit,kim2024switchlight,zhang2025scaling}. Beyond general-purpose relighting, specialized settings have also been studied, including portrait performance relighting with conditional video diffusion and hybrid datasets \cite{mei2025luxpostfactolearning}, as well as approaches that jointly model intrinsic decomposition and relighting synthesis to better capture complex illumination effects \cite{he2025unirelightlearningjointdecomposition}. Among recent methods, IC-Light \cite{zhang2025scaling} demonstrates strong results through scalable diffusion training coupled with light-transport consistency. Building on this line, RelightVid \cite{ponglertnapakorn2024relight} extends relighting to videos via temporal attention, whereas Light-A-Video \cite{zhou2025light} provides a training-free alternative by injecting cross-frame attention during inference.

\textbf{4D Video Generation.} Camera-controllable 4D video generation aims to synthesize temporally coherent dynamic scenes from novel viewpoints.
Existing approaches can be roughly organized by how they enforce 3D/4D structures during generation.
One line of work injects explicit geometric priors or relies on intermediate reconstructions to stabilize view changes.
For instance, EX-4D uses watertight mesh reasoning to support extreme viewpoint variations~\cite{hu2025ex}, and Light-X introduces disentangled geometry cues to enable joint camera and illumination control~\cite{liu2025light}.
Another line emphasizes feed-forward generative modeling over dynamic 3D/4D representations for novel-view rendering, including point-cloud-based formulations in 4DNeX~\cite{chen20254dnex} and diffusion-based 4D occupancy generation in DynamicCity~\cite{bian2025dynamiccity}.
In parallel, camera-conditioned diffusion and pseudo-4D guidance have improved controllability for view synthesis and camera-aware video generation~\cite{fan2025omniview,bian2025gsdit}.

\textbf{4D Relighting.} Relighting in dynamic scenes with viewpoint changes remains relatively underexplored. Most existing systems are training-based and rely on explicit scene representations. Relighting4D \cite{chen2022relighting4d} reconstructs space--time neural fields and uses physically-based rendering to decompose scenes into components such as normals, occlusion, and diffuse effects. Relightable Neural Actor \cite{luvizon2024relightable_neural_actor} targets human performance capture by leveraging multi-view inputs together with intrinsic and material decomposition for pose-conditioned relighting. Light-X \cite{liu2025light} provides a unified pipeline for joint camera and illumination control by training on synthetic paired data, producing videos with coupled viewpoint and lighting variations. BEAM \cite{hong2025beam} represents dynamic scenes using 4D Gaussians and physically-based rendering to recover material properties for high-quality relightable video. Recent monocular relightable Gaussian methods \cite{choi2025rndavatar,jiang2025dnfavatar,schmidt2025becominglit} aim to reduce the need for multi-view capture.

\begin{figure}[ht]
\centering
\includegraphics[width=\columnwidth]{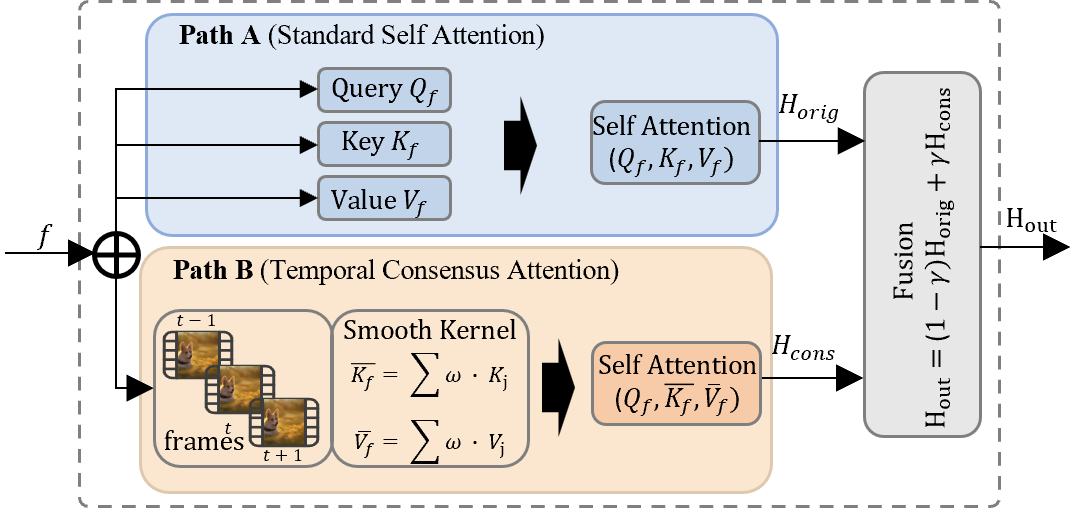}
\caption{\textbf{Design of Temporal Consistent Attention (TCA)}. TCA enforces temporal coherence through a dual-path mechanism that interpolates between standard self-attention (Path A) for frame-specific structure and a consistent path (Path B) that regularizes appearance context via Gaussian-weighted sliding window aggregation.}
\label{fig:light4d_framework}
\vskip -0.2in
\end{figure}

\section{The Proposed Method}
\subsection{Overview}
We propose Light4D, a novel training-free framework designed to synthesize a target 4D video $\mathcal{V} = \{I^f\}_{f=1}^F$ from a monocular source video $\mathcal{V}_s = \{I_s^f\}_{f=1}^F$, as illustrated in Figure~\ref{fig:qualitative}. Our primary objective is to rerender the dynamic scene under a target camera trajectory $\mathcal{C} = \{P_f\}_{f=1}^F$ and user-specified illumination conditions $\mathcal{L}$. Specifically, the camera trajectory $\mathcal{C}$ spans a wide viewpoint range ($-90^{\circ}$ to $90^{\circ}$) relative to the original coordinate system, while the illumination $\mathcal{L}$ is modulated by text prompts corresponding to distinct lighting directions (e.g., Left, Right, Top, Bottom). The generated video $\mathcal{V}$ must faithfully preserve the 3D geometry and motion dynamics of the source $\mathcal{V}_s$, while accurately adhering to the novel viewpoint $\mathcal{C}$ and lighting constraints $\mathcal{L}$. The overall inference pipeline achieving these goals is summarized in Algorithm \ref{alg:inference}.

\subsection{Disentangled Flow Guidance}
\label{sec:3.2}
A critical challenge in 4D relighting arises from the inherent conflict between geometric reconstruction and illumination synthesis within the generative manifold. Standard flow matching models typically prioritize early denoising stages to establish coherent 3D structures. Consequently, the premature injection of high-frequency illumination cues can disrupt this delicate structural formation, resulting in geometric collapse. 

To mitigate these issues, we employ a time-aware strategy that progressively integrates illumination cues into the latent space. Formally, let $z_t$ denote the latent state at timestep $t \in [0, 1]$. We formulate the generative process as guiding a pre-trained geometric flow $\mathbf{v}_{geo}$ instantiated by EX-4D, using an illumination-aware correction derived from a single-view relighting prior $\mathcal{M}_{light}$ based on IC-Light.

At each timestep $t$, we first estimate the clean geometric state $\hat{z}_0^{geo}$ derived from the current flow velocity. Since the relighting prior $\mathcal{M}_{light}$ operates in the image domain, we project this latent estimate into the pixel space via the VAE decoder $\mathcal{D}$. Then, we utilize the relighting prior $\mathcal{M}_{light}$ to predict the relighted image $\hat{x}^{light}_0$ conditioned on the lighting prompt $\mathcal{L}$:
\begin{equation}
\hat{x}_0^{light} = \mathcal{M}_{light}\left(\mathcal{D}(\hat{z}_0^{geo}), \mathcal{L}, \epsilon_{shared}\right),
\label{eq:x0_light}
\end{equation}
where $\epsilon_{shared}$ represents a canonical noise prior.

We then construct a hybrid flow target $z_{target}$ by fusing the geometric structure and illumination cues in the latent space, governed by a time-dependent fusion weight $\lambda(t)$:
\begin{equation}
z_{target} = \mathcal{E}\left( (1 - \lambda(t)) \cdot \mathcal{D}(\hat{z}_0^{geo}) + \lambda(t) \cdot \hat{x}_0^{light} \right),
\label{eq:z_target}
\end{equation}
where $\mathcal{E}$ denotes the VAE encoder. 

This hybrid target $z_{target}$ serves as the target, defining a rectified flow trajectory toward the illumination-enhanced manifold. Based on this target, we perform a discrete ODE update step to compute the state $z_{t'}$ for the next timestep $t'$. Specifically, we employ a first-order Euler solver:
\begin{equation}
z_{t'} = z_t + (\sigma_{t'} - \sigma_t) \cdot \frac{z_t - z_{target}}{\sigma_t + \delta},
\label{eq:euler_update}
\end{equation}
where $\sigma_{t'}$ and $\sigma_t$ denote the noise levels, and $\delta$ is a numerical stabilizer.

Crucially, to mitigate interference between geometry generation and illumination refinement, $\lambda(t)$ follows a multi-phase adaptive schedule. Unlike standard linear annealing, this schedule disentangles the trajectory into four distinct phases, parameterized by the peak intensity $\lambda_{max}$ and the terminal weight $\lambda_{end}$:
\begin{equation}
\lambda(t) =
\begin{cases}
0, & t \in (\tau_g, 1] \\
\lambda_{max} \cdot \sqrt{\frac{\tau_g - t}{\tau_g - \tau_r}}, & t \in [\tau_r, \tau_g] \\
\lambda_{max}, & t \in [\tau_s, \tau_r) \\
\frac{\lambda_{max} - \lambda_{end}}{\tau_s} \cdot t + \lambda_{end}, & t \in [0, \tau_s)
\end{cases}
\end{equation}
where $\tau_g, \tau_r, \tau_s$ are the time thresholds for the geometric, ramp-up, and stable phases, respectively. 

Initially, during the geometric isolation phase ($t > \tau_g$), we enforce $\lambda(t) = 0$ to strictly prohibit illumination cues from interfering with the completion of invisible regions and occlusion relationships. Subsequently, a root-squared ramp $[\tau_r, \tau_g]$ is employed to rapidly anchor lighting conditions without inducing feature shock, followed by a stable plateau $[\tau_s, \tau_r)$ at $\lambda_{max}$ that allows the model to reconcile appearance cues with the established 3D structure. Finally, for $t < \tau_s$, the weight linearly decays to $\lambda_{end}$, ensuring a soft landing that prevents the relighting prior from dominating fine-grained details during convergence.

\subsection{Temporal Consistent Attention}
While Disentangled Flow Guidance (\cref{sec:3.2}) effectively steers the geometric trajectory, the reliance on a frame-independent relighting prior inevitably introduces stochastic fluctuations in the feature space, leading to both geometric instability and temporal flickering. To resolve this, we introduce Temporal Consistent Attention (TCA). As illustrated in Figure~\ref{fig:light4d_framework}, this dual-path mechanism modifies standard self-attention to enforce strict temporal coherence while preserving high-fidelity illumination details.

We redesign its internal self-attention mechanism to assign distinct temporal behaviors. Specifically, the Query features ($\mathbf{Q}$), which encode frame-specific geometric structure, are preserved in their transient state to faithfully capture dynamic changes. In contrast, the Key ($\mathbf{K}$) and Value ($\mathbf{V}$) pairs, which provide the appearance context, are regularized to exhibit strong temporal stability. We employ a Gaussian-weighted sliding window approach to construct a smoothed context $\{\overline{\mathbf{K}}_f, \overline{\mathbf{V}}_f\}$ by aggregating features within a local temporal neighborhood $\mathcal{N}(f) = \{j \mid |f-j| \le r\}$, weighted by their temporal proximity:
\begin{equation}
\begin{aligned}
    \overline{\mathbf{K}}_f &= \sum_{j \in \mathcal{N}(f)} w(|f-j|) \cdot \mathbf{K}_j, \\
    \overline{\mathbf{V}}_f &= \sum_{j \in \mathcal{N}(f)} w(|f-j|) \cdot \mathbf{V}_j,
    \end{aligned}
\end{equation}
where $r$ denotes the temporal radius, $f$ represents the current frame index, and $w(d) = \exp(-d^2 / 2\sigma^2)$ serves as a Gaussian weighting kernel that prioritizes contributions from temporally adjacent frames.

To balance single-frame sharpness with temporal stability, TCA implements a dual-path residual injection strategy. We simultaneously compute the original stochastic features $\mathbf{H}_{\text{orig}}$ using standard self-attention to retain high-frequency details, and the temporally consistent features $\mathbf{H}_{\text{cons}}$ by attending the frame-specific query $\mathbf{Q}_f$ to the temporally smoothed context $\{\overline{\mathbf{K}}_f, \overline{\mathbf{V}}_f\}$:
\begin{equation}
\begin{aligned}
    \mathbf{H}_{\text{orig}} &= \text{Attention}(\mathbf{Q}_f, \mathbf{K}_f, \mathbf{V}_f), \\
    \mathbf{H}_{\text{cons}} & = \text{Attention}(\mathbf{Q}_f, \overline{\mathbf{K}}_f, \overline{\mathbf{V}}_f).
    \end{aligned}
\end{equation}
Subsequently, the final hidden state is derived via a linear interpolation controlled by a mixing coefficient $\gamma \in [0, 1]$:
\begin{equation}
    \mathbf{H}_{out} = (1 - \gamma) \mathbf{H}_{\text{orig}} + \gamma \mathbf{H}_{\text{cons}}.
\end{equation}
This residual formulation effectively mitigates the propagation of temporal variance into the rectification target while preserving the structural distinctiveness of individual frames.

\begin{figure*}[ht]
\centering
\includegraphics[width=\textwidth]{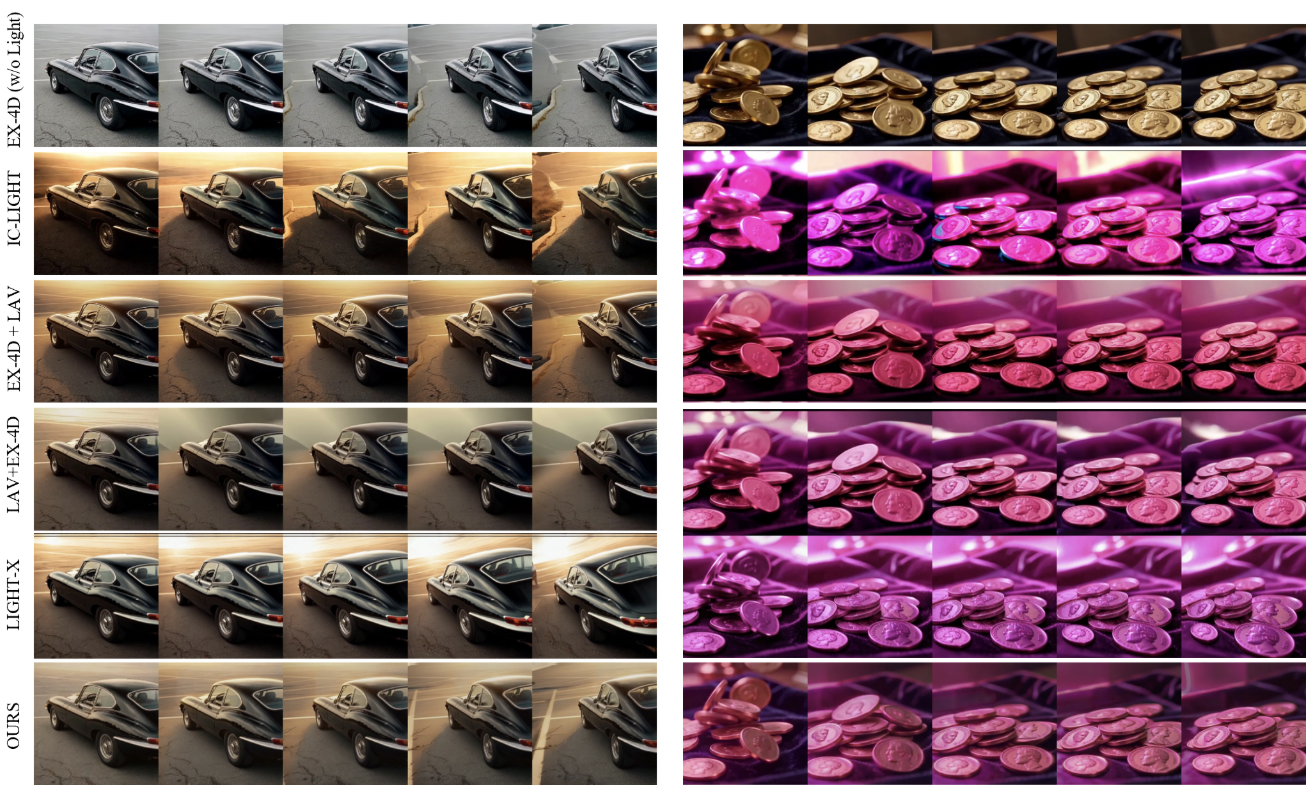}
\caption{\textbf{Qualitative relighting results.} Comparison of baselines and our method under two prompts: ``Sunlight'' (left) and ``Pink neon light'' (right). Our method yields more stable illumination changes over time and reduces temporal flicker.}
\label{fig:relighting_vis}
\vskip -0.2in
\end{figure*}

\begin{table*}[ht]
\caption{\textbf{Relighting-related metrics across viewpoint changes.}
We report CLIP-Frame, Motion Flow L1, HFPR, and Aesthetic Score under camera motion ranges of $30^{\circ}$, $90^{\circ}$, and $180^{\circ}$.
These metrics jointly reflect the trade-off between \emph{lighting consistency}, \emph{4D geometric and motion stability}, and \emph{detail fidelity} under extreme viewpoints.
Best results are \textbf{bolded} and second-best are \underline{underlined}.}
\label{tab:relighting}
\centering
\begin{small}
\begin{sc}
\setlength{\tabcolsep}{2.5pt}
\renewcommand{\arraystretch}{0.95}
\begin{tabular*}{\textwidth}{l @{\extracolsep{\fill}}cccccccccccc}
\toprule
\multirow{2}{*}{\textbf{Method}} & \multicolumn{3}{c}{\textbf{CLIP-Frame} $\uparrow$} & \multicolumn{3}{c}{\textbf{Motion Flow L1} $\downarrow$} & \multicolumn{3}{c}{\textbf{HFPR} $\uparrow$} & \multicolumn{3}{c}{\textbf{Aesthetic} $\uparrow$} \\
\cmidrule(lr){2-4} \cmidrule(lr){5-7} \cmidrule(lr){8-10} \cmidrule(lr){11-13}
 & $30^{\circ}$ & $90^{\circ}$ & $180^{\circ}$ & $30^{\circ}$ & $90^{\circ}$ & $180^{\circ}$ & $30^{\circ}$ & $90^{\circ}$ & $180^{\circ}$ & $30^{\circ}$ & $90^{\circ}$ & $180^{\circ}$ \\
\midrule
\textit{Training-based} & & & & & & & & & & & & \\
Light-X & 0.956 & 0.900 & 0.901 & \underline{0.814}& \textbf{1.832} & \underline{4.470} & 0.945 & 0.949 & 0.931 & \underline{0.235} & \underline{0.229} & \underline{0.221} \\
\midrule
\textit{Training-free} & & & & & & & & & & & & \\
EX-4D + IC-Light & 0.923 & 0.885 & 0.885 & 15.725 & 21.652 & 24.114 & 0.842 & 0.838 & 0.817 & 0.228 & 0.223 & 0.219 \\
EX-4D + LAV & \underline{0.961} & \underline{0.925} & \underline{0.923} & 1.287 & 3.200 & 6.095 & \underline{0.951} & \underline{0.959} & \underline{0.946} & 0.209 & 0.189 & 0.168 \\
LAV + EX-4D & 0.959 & 0.918 & 0.922 & \underline{0.982} & \underline{2.923} & 4.476 & 0.949 & 0.932 & 0.944 & 0.199 & 0.174 & 0.169 \\
\midrule
\textbf{Light4D (Ours)} & \textbf{0.975}& \textbf{0.930} & \textbf{0.930} & \textbf{0.791} & \underline{1.936} & \textbf{3.524} & \textbf{0.974} & \textbf{0.963} & \textbf{0.966} & \textbf{0.243} & \textbf{0.235} & \textbf{0.231} \\
\bottomrule
\end{tabular*}
\end{sc}
\end{small}
\vskip -0.1in
\end{table*}

\begin{table*}[ht]
\caption{\textbf{Video quality metrics across viewpoint changes.}
We measure frame-wise reconstruction similarity between the relighting videos and the corresponding source videos under viewpoint changes of $30^{\circ}$, $90^{\circ}$, and $180^{\circ}$.
We report per-frame PSNR, per-frame SSIM, and per-frame LPIPS to quantify content and detail preservation.
Best results are \textbf{bolded} and second-best are \underline{underlined}.}
\label{tab:video_quality}
\centering
\begin{small}
\begin{sc}
\setlength{\tabcolsep}{2.5pt}
\renewcommand{\arraystretch}{0.95}
\begin{tabular*}{\textwidth}{l @{\extracolsep{\fill}}ccccccccc}
\toprule
\multirow{2}{*}{\textbf{Method}} & \multicolumn{3}{c}{\textbf{Frame PSNR} $\uparrow$} & \multicolumn{3}{c}{\textbf{Frame SSIM} $\uparrow$} & \multicolumn{3}{c}{\textbf{Frame LPIPS} $\downarrow$} \\
\cmidrule(lr){2-4} \cmidrule(lr){5-7} \cmidrule(lr){8-10}
 & $30^{\circ}$ & $90^{\circ}$ & $180^{\circ}$ & $30^{\circ}$ & $90^{\circ}$ & $180^{\circ}$ & $30^{\circ}$ & $90^{\circ}$ & $180^{\circ}$ \\
\midrule
\textit{Training-based} & & & & & & & & & \\
Light-X & \underline{12.899} & \underline{13.544} & \underline{13.831} & \underline{0.738} & \underline{0.733} & \underline{0.752} & \textbf{0.349} & \underline{0.358} & \underline{0.332} \\
\midrule
\textit{Training-free} & & & & & & & & & \\
EX-4D + IC-Light & 9.147 & 9.412 & 9.035 & 0.493 & 0.489 & 0.473 & 0.754 & 0.760 & 0.748 \\
EX-4D + LAV & 11.957 & 11.843 & 11.449 & 0.696 & 0.659 & 0.652 & 0.572 & 0.625 & 0.624 \\
LAV + EX-4D & 12.348 & 12.007 & 12.023 & 0.636 & 0.564 & 0.554 & 0.535 & 0.560 & 0.552 \\
\midrule
\textbf{Light4D (Ours)} & \textbf{14.056} & \textbf{13.728} & \textbf{13.941} & \textbf{0.761} & \textbf{0.759} & \textbf{0.753} & \underline{0.360} & \textbf{0.341} & \textbf{0.307} \\
\bottomrule
\end{tabular*}%
\end{sc}
\end{small}
\vskip -0.1in
\end{table*}

\begin{table}[ht]
\caption{\textbf{Ablation Study: Relighting Metrics at $30^{\circ}$.} Best results are \textbf{bolded}; second best are \underline{underlined}.}
\label{tab:ablation_relighting_30}
\centering
\begin{small}
\begin{sc}
\renewcommand{\arraystretch}{0.95}
\begin{tabular*}{\columnwidth}{l @{\extracolsep{\fill}} c c c c}
\toprule
Method & CLIP $\uparrow$ & Motion $\downarrow$ & HFPR $\uparrow$ & Aes $\uparrow$ \\
\midrule
w/o CLA & 0.972& 0.903 & 0.882& 0.229\\
w/o DGA & 0.965& 1.259 & 0.754 & 0.213\\
\midrule
w/o FDI & 0.970& 0.885& 0.910& 0.228\\
w/o CNI & 0.971& 0.875& 0.925& 0.229\\
w/o GMM & 0.972& 0.860& 0.940& 0.230\\
w/o All & \underline{0.793}& \underline{0.898} & \underline{0.893}& \underline{0.231}\\
\midrule
\textbf{Full Model} & \textbf{0.975}& \textbf{0.791}& \textbf{0.974}& \textbf{0.243}\\
\bottomrule
\end{tabular*}
\end{sc}
\end{small}
\vskip -0.1in
\end{table}

\textbf{Deterministic Coherence.}
Before fusing the IC-Light prediction into the hybrid target, we apply a deterministic stabilization procedure to suppress temporal stochasticity, including Canonical Noise Initialization (CNI), Global Moment Matching (GMM), and Frequency-Decoupled Illuminance Regularization (FDI). More details are provided in \cref{app:det_coh}.

\begin{figure*}[t]
\begin{center}
\includegraphics[width=\textwidth]{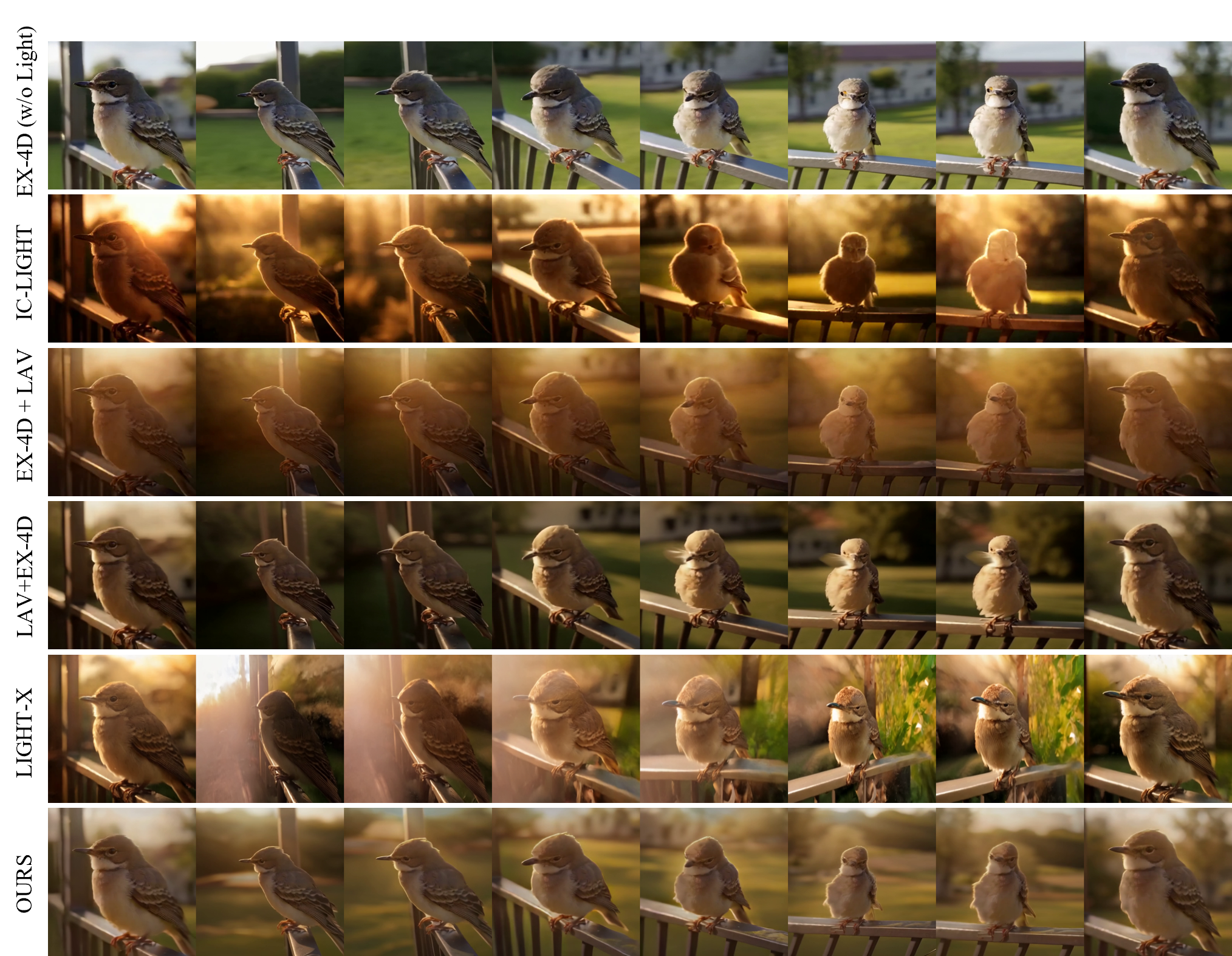}
\caption{\textbf{4D video quality under extreme viewpoints.}
Qualitative comparison of baselines and our method using the prompt ``Sunlight''.
Under extreme viewpoint changes, our method better balances relighting coherence, 4D geometric stability, and detail fidelity.}
\label{fig:video_quality_vis}
\end{center}
\vskip -0.2in
\end{figure*}

\section{Experiment}

\subsection{Experimental Setup}

\textbf{Baselines.}
Since Light4D is the first training-free framework for 4D video relighting, we compare it with both state-of-the-art training-based methods and cascaded training-free baselines. For training-based approaches, we select Light-X~\cite{liu2025light}, which is trained to jointly control camera motion and illumination.
For training-free comparisons, we construct two sequential pipelines: EX-4D~\cite{hu2025ex} $\rightarrow$ Light-A-Video (LAV)~\cite{zhou2025light}, and LAV~\cite{zhou2025light} $\rightarrow$ EX-4D~\cite{hu2025ex}.
We also include a naive baseline EX-4D~\cite{hu2025ex} + IC-Light~\cite{zhang2025scaling} to expose a lower bound on temporal stability when relighting is applied independently to each frame.

\textbf{Benchmark.}
To evaluate 4D relighting under extreme viewpoint changes, we generated an evaluation benchmark of 100 high-quality videos produced by recent video models (Sora~\cite{videoworldsimulators2024}, WanVideo~\cite{wan2025wan}, and Kling~\cite{klingteam2025klingomnitechnicalreport}), spanning humans, animals, objects, and landscapes.
Each video is evaluated in three movement ranges of the camera ($30^{\circ}, 90^{\circ}, 180^{\circ}$).
To investigate generalization to real captured videos, we additionally apply our method to real-world driving sequences from the OpenScene dataset~\cite{OpenScene_Dataset_Contributors_OpenScene_The_Largest_2023} and use these results to qualitatively analyze lighting behavior on real videos.

\textbf{Evaluation Metrics.}
We evaluate relit videos against source videos under three camera movement ranges ($30^{\circ}$, $90^{\circ}$, $180^{\circ}$) using two metric groups.
\emph{Relighting-related} metrics include:
(i) \emph{CLIP-Frame}~\cite{radford2021clip}, the inter-frame similarity of CLIP embeddings computed on consecutive frames to quantify temporal coherence;
(ii) \emph{Motion Flow L1}, the mean distance $\ell_1$ between RAFT~\cite{teed2020raft} optical flows estimated from relit and source videos to measure motion preservation;
(iii) \emph{High Frequency Preservation Ratio (HFPR)}~\cite{zhu2012tpr,zhu2013laplacianvqa}, the ratio of Laplacian-filtered high-frequency energy between the relit and source videos to assess detail retention; and
(iv) \emph{Aesthetic Score}~\cite{schuhmann2022laion} for perceptual appeal.
\emph{Video quality} metrics measure reconstruction fidelity:
(i) \emph{Frame PSNR}, and (ii--iii) \emph{warp-aligned} SSIM and LPIPS, where we estimate flow using RAFT and warp adjacent frames to a common reference before computing the metrics, thereby reducing motion-induced misalignment and better isolating appearance and structural consistency over time~\cite{wang2004ssim,zhang2018lpips}.

\textbf{Implementation Details.}
Our framework is based on EX-4D~\cite{hu2025ex} as the generative backbone of 4D and IC-Light~\cite{zhang2025scaling} as the illumination prior.
Our default setting generates $49$-frame videos at a resolution of $384\times384$ with $T=25$ denoising steps, and all experiments are run on NVIDIA H20 GPU.
To mitigate geometry–illumination interference, we adopt a time-aware fusion strategy that delays illumination injection to later denoising stages, prioritizing 4D structure formation before enforcing relighting cues.
We further improve temporal stability via a lightweight temporal consistency module and post-processing smoothing. The complete hyperparameters and trajectory alignment details are provided in the Appendix~\ref{app:impl_details}.

\subsection{Main Results}

\textbf{Relighting Quality Results.}
We visualize relighting outputs under two lighting conditions, ``Sunlight'' (left) and ``Pink neon light'' (right), as shown in Figure~\ref{fig:relighting_vis}.
Quantitatively (Table~\ref{tab:relighting}), our method achieves the strongest overall performance at $30^{\circ}$, $90^{\circ}$, and $180^{\circ}$ in CLIP-Frame, HFPR, and Aesthetic Score, suggesting improved temporal coherence and perceptual quality while preserving fine details.
These gains are consistent with the visual comparisons in Figure~\ref{fig:relighting_vis}, where our relighting exhibits smoother light transitions and fewer flickering artifacts.
The naive EX-4D+IC-Light baseline~\cite{hu2025ex,zhang2025scaling} can yield sharp individual frames, but frame-wise relighting introduces severe temporal inconsistency, resulting in higher Motion Flow L1 and visibly unstable illumination.
The cascade pipelines exhibit different trade-offs: both EX-4D$\rightarrow$LAV and LAV$\rightarrow$EX-4D~\cite{hu2025ex,zhou2025light} improve temporal behavior compared to frame-wise relighting; however, yet their lighting cues are not consistently aligned with the evolving 4D geometry under large rotations. This often leads to less natural appearance and reduced motion or geometry stability as the viewpoint change increases.
Light-X~\cite{liu2025light} performs well at moderate rotations, but degrades in larger viewpoints settings. In particular, visualizations reveal a more static, “baked-in” illumination pattern that does not adapt faithfully when newly visible geometry emerges.

\textbf{4D Video Quality Results.}
Relighting 4D content while preserving geometry and fine-grained content under extreme viewpoints is particularly challenging.
Figure~\ref{fig:video_quality_vis} shows qualitative results under the ``Sunlight'' prompt, and Table~\ref{tab:video_quality} reports frame-wise reconstruction metrics between relit and source videos.
Our method consistently leads in Frame PSNR and SSIM across all viewpoint ranges and achieves the lowest LPIPS at $90^{\circ}$ and $180^{\circ}$, indicating strong content preservation without sacrificing perceptual detail.
This indicates that Light4D preserves the underlying 4D content while applying substantial illumination edits, even as viewpoint changes increase.
This indicates that Light4D preserves the underlying 4D content while applying substantial illumination edits, even as viewpoint changes increase.
In contrast, cascaded baselines~\cite{hu2025ex, zhou2025light, zhang2025scaling} are more sensitive to large viewpoint changes: introducing relighting cues either before or after 4D generation can amplify small geometric inconsistencies into visible distortions when rotations reach $90^{\circ}$ and beyond.
Light-X~\cite{liu2025light} maintains reasonable fidelity, yet shows marginally lower PSNR and SSIM scores in this extreme-view evaluation.
In summary, our method offers a more balanced compromise under extreme camera motion, simultaneously maintaining illumination coherence, stabilizing 4D geometry, and preserving video details.
\subsection{Ablation Studies}
We validate the effectiveness of our design by ablating key components and reporting results from a $30^{\circ}$ viewpoint. Specifically, we isolate the impact of each proposed module to demonstrate its necessity in our pipeline. Table~\ref{tab:ablation_relighting_30} summarizes relighting metrics, while the full set of metrics (Table~\ref{tab:ablation_relighting_30_app}) and more ablation details are provided in Appendix~\ref{app:abla details}.

\textbf{Impact of Disentangled Guidance (DGA).}
Removing DGA (\emph{w/o DGA}) markedly degrades relighting quality: Motion Flow L1 increases substantially and HFPR drops, indicating weaker motion preservation and loss of fine details. This supports the role of DGA in injecting illumination cues without destabilizing the underlying 4D geometry.

\textbf{Effect of Consistent Light Attention (CLA).}
Disabling CLA (\emph{w/o CLA}) reduces temporal coherence in relighting, reflected by higher Motion Flow L1 compared to the full model. It also leads to more noticeable frame-to-frame lighting jitter in visually smooth regions. This suggests that CLA is important for stabilizing illumination across consecutive frames.

\textbf{Effect of Regularization.}
We further ablate the Canonical Noise Initialization (CNI), Global Moment Matching (GMM), and Frequency-Decoupled Illuminance (FDI). Overall, removing these regularizers harms relighting quality in complementary ways. Specifically, removing all smoothing increases temporal instability, whereas the full model preserves sharp details with higher HFPR and achieves better perceptual quality with a higher Aesthetic Score. Across our regularization modules, we observe the largest degradation when removing FDI, followed by CNI, and then GMM. While their effects are partially complementary and can vary slightly across metrics, the overall trend is consistent across temporal coherence and detail-preservation measures.

\section{Limitation and Future Work}
As a training-free framework, Light4D's performance is bounded by the capabilities of its foundation models, specifically the geometric fidelity of EX-4D and the single-frame nature of the IC-Light prior. Consequently, maintaining rigorous global lighting consistency remains challenging when adapting 2D image priors to the temporal domain, particularly during extreme viewpoint traversals ($-90^{\circ}\sim90^{\circ}$). Future work will leverage the framework's modularity to integrate more advanced video-native illumination models and stronger 4D generative backbones to further improve synthesis quality. We also aim to investigate specialized mechanisms to mitigate photometric inconsistencies induced by large viewpoint variations, while extending the framework to handle complex light transport effects, such as cast shadows and inter-reflections.

\section{Conclusion}

In this paper, we present Light4D, a training-free framework for 4D video relighting under extreme viewpoint changes. By identifying and resolving the conflict between geometric reconstruction and illumination synthesis, we propose Disentangled Flow Guidance to harmonize these objectives with a time-aware schedule. Furthermore, we introduce Temporal Consistent Attention and Deterministic Coherence Regularization to ensure the generated content remains geometrically consistent and free of visible flicker. Extensive experiments show that our method achieves competitive performance in both geometric consistency and lighting fidelity compared to baselines, offering a scalable solution for controllable 4D content creation. Finally, the modular design of our framework makes it  extensible, enabling integration of future advances in generative models.

\FloatBarrier

\bibliography{example_paper}

@inproceedings{bahmani20244d,
  title={4d-fy: Text-to-4d generation using hybrid score distillation sampling},
  author={Bahmani, Sherwin and Skorokhodov, Ivan and Rong, Victor and Wetzstein, Gordon and Guibas, Leonidas and Wonka, Peter and Tulyakov, Sergey and Park, Jeong Joon and Tagliasacchi, Andrea and Lindell, David B},
  booktitle={Proceedings of the IEEE/CVF Conference on Computer Vision and Pattern Recognition},
  pages={7996--8006},
  year={2024}
}

@inproceedings{pang2025disco4d,
  title={Disco4d: Disentangled 4d human generation and animation from a single image},
  author={Pang, Hui En and Liu, Shuai and Cai, Zhongang and Yang, Lei and Zhang, Tianwei and Liu, Ziwei},
  booktitle={Proceedings of the Computer Vision and Pattern Recognition Conference},
  pages={26331--26344},
  year={2025}
}

@article{wen2025dynamicverse,
  title={DynamicVerse: A Physically-Aware Multimodal Framework for 4D World Modeling},
  author={Wen, Kairun and Huang, Yuzhi and Chen, Runyu and Zheng, Hui and Lin, Yunlong and Pan, Panwang and Li, Chenxin and Cong, Wenyan and Zhang, Jian and Lu, Junbin and others},
  journal={arXiv preprint arXiv:2512.03000},
  year={2025}
}

@article{lin2025illumicraft,
  title={IllumiCraft: Unified Geometry and Illumination Diffusion for Controllable Video Generation},
  author={Lin, Yuanze and Chen, Yi-Wen and Tsai, Yi-Hsuan and Clark, Ronald and Yang, Ming-Hsuan},
  journal={arXiv preprint arXiv:2506.03150},
  year={2025}
}

@inproceedings{bharadwaj2025genlit,
  title={GenLit: Reformulating single-image relighting as video generation},
  author={Bharadwaj, Shrisha and Feng, Haiwen and Becherini, Giorgio and Fernandez Abrevaya, Victoria and Black, Michael J},
  booktitle={Proceedings of the SIGGRAPH Asia 2025 Conference Papers},
  pages={1--12},
  year={2025}
}

@article{chen20254dnex,
  title={4dnex: Feed-forward 4d generative modeling made easy},
  author={Chen, Zhaoxi and Liu, Tianqi and Zhuo, Long and Ren, Jiawei and Tao, Zeng and Zhu, He and Hong, Fangzhou and Pan, Liang and Liu, Ziwei},
  journal={arXiv preprint arXiv:2508.13154},
  year={2025}
}

@article{mi2025one4d,
  title={One4D: Unified 4D Generation and Reconstruction via Decoupled LoRA Control},
  author={Mi, Zhenxing and Wang, Yuxin and Xu, Dan},
  journal={arXiv preprint arXiv:2511.18922},
  year={2025}
}

@inproceedings{zhou2025holotime,
  title={Holotime: Taming video diffusion models for panoramic 4d scene generation},
  author={Zhou, Haiyang and Yu, Wangbo and Guan, Jiawen and Cheng, Xinhua and Tian, Yonghong and Yuan, Li},
  booktitle={Proceedings of the 33rd ACM International Conference on Multimedia},
  pages={9763--9772},
  year={2025}
}

@article{zhang2021nerfactor,
  title={Nerfactor: Neural factorization of shape and reflectance under an unknown illumination},
  author={Zhang, Xiuming and Srinivasan, Pratul P and Deng, Boyang and Debevec, Paul and Freeman, William T and Barron, Jonathan T},
  journal={ACM Transactions on Graphics (ToG)},
  volume={40},
  number={6},
  pages={1--18},
  year={2021},
  publisher={ACM New York, NY, USA}
}

@inproceedings{chaturvedi2025synthlight,
  title={SynthLight: Portrait Relighting with Diffusion Model by Learning to Re-render Synthetic Faces},
  author={Chaturvedi, Sumit and Ren, Mengwei and Hold-Geoffroy, Yannick and Liu, Jingyuan and Dorsey, Julie and Shu, Zhixin},
  booktitle={Proceedings of the Computer Vision and Pattern Recognition Conference},
  pages={369--379},
  year={2025}
}

@article{hu2025ex,
  title={EX-4D: EXtreme Viewpoint 4D Video Synthesis via Depth Watertight Mesh},
  author={Hu, Tao and Peng, Haoyang and Liu, Xiao and Ma, Yuewen},
  journal={arXiv preprint arXiv:2506.05554},
  year={2025}
}

@inproceedings{zhang2025scaling,
  title={Scaling in-the-wild training for diffusion-based illumination harmonization and editing by imposing consistent light transport},
  author={Zhang, Lvmin and Rao, Anyi and Agrawala, Maneesh},
  booktitle={The Thirteenth International Conference on Learning Representations},
  year={2025}
}

@article{liu2025light,
  title={Light-X: Generative 4D Video Rendering with Camera and Illumination Control},
  author={Liu, Tianqi and Chen, Zhaoxi and Huang, Zihao and Xu, Shaocong and Zhang, Saining and Ye, Chongjie and Li, Bohan and Cao, Zhiguo and Li, Wei and Zhao, Hao and others},
  journal={arXiv preprint arXiv:2512.05115},
  year={2025}
}

@inproceedings{wang2025comprehensive_relighting,
  title={Comprehensive Relighting: Generalizable and Consistent Monocular Human Relighting and Harmonization},
  author={Wang, Junying and Liu, Jingyuan and Sun, Xin and Singh, Krishna Kumar and Shu, Zhixin and Zhang, He and Yang, Jimei and Zhao, Nanxuan and Wang, Tuanfeng Y and Chen, Simon S and others},
  booktitle={Proceedings of the Computer Vision and Pattern Recognition Conference},
  pages={380--390},
  year={2025}
}

@article{zhou2025light,
  title={Light-a-video: Training-free video relighting via progressive light fusion},
  author={Zhou, Yujie and Bu, Jiazi and Ling, Pengyang and Zhang, Pan and Wu, Tong and Huang, Qidong and Li, Jinsong and Dong, Xiaoyi and Zang, Yuhang and Cao, Yuhang and others},
  journal={arXiv preprint arXiv:2502.08590},
  year={2025}
}

@misc{he2025unirelightlearningjointdecomposition,
  title={UniRelight: Learning Joint Decomposition and Synthesis for Video Relighting},
  author={Kai He and Ruofan Liang and Jacob Munkberg and Jon Hasselgren and Nandita Vijaykumar and Alexander Keller and Sanja Fidler and Igor Gilitschenski and Zan Gojcic and Zian Wang},
  year={2025},
  eprint={2506.15673},
  archivePrefix={arXiv},
  primaryClass={cs.CV},
  url={https://arxiv.org/abs/2506.15673}
}

@misc{mei2025luxpostfactolearning,
  title={Lux Post Facto: Learning Portrait Performance Relighting with Conditional Video Diffusion and a Hybrid Dataset},
  author={Yiqun Mei and Mingming He and Li Ma and Julien Philip and Wenqi Xian and David M George and Xueming Yu and Gabriel Dedic and Ahmet Levent Taşel and Ning Yu and Vishal M. Patel and Paul Debevec},
  year={2025},
  eprint={2503.14485},
  archivePrefix={arXiv},
  primaryClass={cs.GR},
  url={https://arxiv.org/abs/2503.14485}
}

@inproceedings{zhang2021neural_video_portrait_relighting,
  title={Neural video portrait relighting in real-time via consistency modeling},
  author={Zhang, Longwen and Zhang, Qixuan and Wu, Minye and Yu, Jingyi and Xu, Lan},
  booktitle={Proceedings of the IEEE/CVF international conference on computer vision},
  pages={802--812},
  year={2021}
}

@misc{chandran2022temporally,
  title={Temporally consistent relighting for portrait videos},
  author={Chandran, Sreenithy and Hold-Geoffroy, Yannick and Sunkavalli, Kalyan and Shu, Zhixin and Jayasuriya, Suren},
  howpublished={In Proceedings of the IEEE/CVF Winter Conference on Applications of Computer Vision},
  pages={719--728},
  year={2022}
}

@misc{ponglertnapakorn2023difareli,
  title={Difareli: Diffusion face relighting},
  author={Ponglertnapakorn, Puntawat and Tritrong, Nontawat and Suwajanakorn, Supasorn},
  booktitle={Proceedings of the IEEE/CVF international conference on computer vision},
  pages={22646--22657},
  year={2023}
}

@article{fang2025relightvid,
  title={RelightVid: Temporal-consistent diffusion model for video relighting},
  author={Fang, Ye and Sun, Zeyi and Zhang, Shangzhan and Wu, Tong and Xu, Yinghao and Zhang, Pan and Wang, Jiaqi and Wetzstein, Gordon and Lin, Dahua},
  journal={arXiv preprint arXiv:2501.16330},
  year={2025}
}

@article{magar2025lightlab,
  title={LightLab: Controlling Light Sources in Images with Diffusion Models},
  author={Magar, Nadav and Hertz, Amir and Tabellion, Eric and Pritch, Yael and Rav-Acha, Alex and Shamir, Ariel and Hoshen, Yedid},
  journal={arXiv preprint arXiv:2505.09608},
  year={2025}
}

@inproceedings{kocsis2024lightit,
  title={Lightit: Illumination modeling and control for diffusion models},
  author={Kocsis, Peter and Philip, Julien and Sunkavalli, Kalyan and Nie{\\ss}ner, Matthias and Hold-Geoffroy, Yannick},
  booktitle={Proceedings of the IEEE/CVF Conference on Computer Vision and Pattern Recognition},
  pages={9359--9369},
  year={2024}
}

@inproceedings{kim2024switchlight,
  title={Switchlight: Co-design of physics-driven architecture and pre-training framework for human portrait relighting},
  author={Kim, Hoon and Jang, Minje and Yoon, Wonjun and Lee, Jisoo and Na, Donghyun and Woo, Sanghyun},
  booktitle={Proceedings of the IEEE/CVF Conference on Computer Vision and Pattern Recognition},
  pages={25096--25106},
  year={2024}
}

@article{ponglertnapakorn2024relight,
  title={DiFaReli++: Diffusion Face Relighting with Consistent Cast Shadows},
  author={Ponglertnapakorn, Puntawat and Tritrong, Nontawat and Suwajanakorn, Supasorn},
  journal={IEEE transactions on pattern analysis and machine intelligence},
  year={2025}
}

@article{bian2025dynamiccity,
  title={Dynamiccity: Large-scale 4d occupancy generation from dynamic scenes},
  author={Bian, Hengwei and Kong, Lingdong and Xie, Haozhe and Pan, Liang and Qiao, Yu and Liu, Ziwei},
  journal={arXiv preprint arXiv:2410.18084},
  year={2024}
}

@article{fan2025omniview,
  title={OmniView: An All-Seeing Diffusion Model for 3D and 4D View Synthesis},
  author={Fan, Xiang and Girish, Sharath and Ramanujan, Vivek and Wang, Chaoyang and Mirzaei, Ashkan and Sushko, Petr and Siarohin, Aliaksandr and Tulyakov, Sergey and Krishna, Ranjay},
  journal={arXiv preprint arXiv:2512.10940},
  year={2025}
}

@inproceedings{bian2025gsdit,
  title={GS-DiT: Advancing Video Generation with Dynamic 3D Gaussian Fields through Efficient Dense 3D Point Tracking},
  author={Bian, Weikang and Huang, Zhaoyang and Shi, Xiaoyu and Li, Yijin and Wang, Fu-Yun and Li, Hongsheng},
  booktitle={Proceedings of the Computer Vision and Pattern Recognition Conference},
  pages={21717--21727},
  year={2025}
}

@misc{chen2022relighting4d,
  title={Relighting4D: Neural Relightable Human from Videos},
  author={Zhaoxi Chen and Ziwei Liu},
  year={2022},
  eprint={2207.07104},
  archivePrefix={arXiv},
  primaryClass={cs.CV},
  url={https://arxiv.org/abs/2207.07104}
}

@misc{luvizon2024relightable_neural_actor,
  title={Relightable Neural Actor with Intrinsic Decomposition and Pose Control},
  author={Diogo Luvizon and Vladislav Golyanik and Adam Kortylewski and Marc Habermann and Christian Theobalt},
  year={2024},
  eprint={2312.11587},
  archivePrefix={arXiv},
  primaryClass={cs.CV},
  url={https://arxiv.org/abs/2312.11587}
}

@article{hong2025beam,
  title={BEAM: Bridging Physically-based Rendering and Gaussian Modeling for Relightable Volumetric Video},
  author={Hong, Yu and Wu, Yize and Shen, Zhehao and Guo, Chengcheng and Jiang, Yuheng and Zhang, Yingliang and Yu, Jingyi and Xu, Lan},
  journal={arXiv preprint arXiv:2502.08297},
  year={2025}
}

@inproceedings{choi2025rndavatar,
  title={Relightable and Dynamic Gaussian Avatar Reconstruction from Monocular Video},
  author={Choi, Seonghwa and Choi, Moonkyeong and Jang, Mingyu and Kim, Jaekyung and Cai, Jianfei and Cheng, Wen-Huang and Lee, Sanghoon},
  booktitle={Proceedings of the 33rd ACM International Conference on Multimedia},
  pages={7405--7414},
  year={2025}
}

@article{jiang2025dnfavatar,
  title={DNF-Avatar: Distilling Neural Fields for Real-time Animatable Avatar Relighting},
  author={Jiang, Zeren and Wang, Shaofei and Tang, Siyu},
  journal={arXiv preprint arXiv:2504.10486},
  year={2025}
}

@article{schmidt2025becominglit,
  title={BecomingLit: Relightable Gaussian Avatars with Hybrid Neural Shading},
  author={Schmidt, Jonathan and Giebenhain, Simon and Niessner, Matthias},
  journal={arXiv preprint arXiv:2506.06271},
  year={2025}
}

@misc{OpenScene_Dataset_Contributors_OpenScene_The_Largest_2023,
author = {{OpenScene Dataset Contributors}},
month = aug,
title = {{OpenScene: The Largest Up-to-Date 3D Occupancy Prediction Benchmark in Autonomous Driving}},
url = {https://github.com/OpenDriveLab/OpenScene},
year = {2023}
}

@inproceedings{zhu2012tpr,
  title        = {An objective method of measuring texture preservation for camcorder performance evaluation},
  author       = {Zhu, Kongfeng and Li, Shujun and Saupe, Dietmar},
  booktitle    = {Proceedings of SPIE-IS\&T Electronic Imaging: Image Quality and System Performance IX},
  volume       = {8293},
  pages        = {829304},
  year         = {2012},
  doi          = {10.1117/12.907265}
}

@inproceedings{zhu2013laplacianvqa,
  title        = {A no-reference video quality assessment based on {Laplacian} pyramids},
  author       = {Zhu, Kongfeng and Hirakawa, Keigo and Asari, Vijayan K. and Saupe, Dietmar},
  booktitle    = {2013 IEEE International Conference on Image Processing (ICIP)},
  pages        = {49--53},
  year         = {2013},
  organization = {IEEE}
}

@inproceedings{radford2021clip,
  title     = {Learning Transferable Visual Models From Natural Language Supervision},
  author    = {Radford, Alec and Kim, Jong Wook and Hallacy, Chris and Ramesh, Aditya and Goh, Gabriel and Agarwal, Sandhini and Sastry, Girish and Askell, Amanda and Mishkin, Pamela and Clark, Jack and Krueger, Gretchen and Sutskever, Ilya},
  booktitle = {Proceedings of the 38th International Conference on Machine Learning (ICML)},
  year      = {2021}
}

@inproceedings{teed2020raft,
  title     = {{RAFT}: Recurrent All-Pairs Field Transforms for Optical Flow},
  author    = {Teed, Zachary and Deng, Jia},
  booktitle = {European Conference on Computer Vision (ECCV)},
  year      = {2020}
}

@inproceedings{schuhmann2022laion,
  title     = {{LAION}-5B: An open large-scale dataset for training next generation image-text models},
  author    = {Schuhmann, Christoph and Beaumont, Romain and Vencu, Richard and Gordon, Cade W and Wightman, Ross and Cherti, Mehdi and Coombes, Theo and Katta, Aarush and Mullis, Clayton and Wortsman, Mitchell and Schramowski, Patrick and Kundurthy, Srivatsa R and Crowson, Katherine and Schmidt, Ludwig and Kaczmarczyk, Robert and Jitsev, Jenia},
  booktitle = {NeurIPS Datasets and Benchmarks},
  year      = {2022}
}

@article{wang2004ssim,
  title   = {Image Quality Assessment: From Error Visibility to Structural Similarity},
  author  = {Wang, Zhou and Bovik, Alan C. and Sheikh, Hamid R. and Simoncelli, Eero P.},
  journal = {IEEE Transactions on Image Processing},
  volume  = {13},
  number  = {4},
  pages   = {600--612},
  year    = {2004}
}

@inproceedings{zhang2018lpips,
  title     = {The Unreasonable Effectiveness of Deep Features as a Perceptual Metric},
  author    = {Zhang, Richard and Isola, Phillip and Efros, Alexei A. and Shechtman, Eli and Wang, Oliver},
  booktitle = {IEEE Conference on Computer Vision and Pattern Recognition (CVPR)},
  year      = {2018}
}

@article{videoworldsimulators2024,
  title={Video generation models as world simulators},
  author={Tim Brooks and Bill Peebles and Connor Holmes and Will DePue and Yufei Guo and Li Jing and David Schnurr and Joe Taylor and Troy Luhman and Eric Luhman and Clarence Ng and Ricky Wang and Aditya Ramesh},
  year={2024},
  url={https://openai.com/research/video-generation-models-as-world-simulators},
}

@article{wan2025wan,
  title={Wan: Open and advanced large-scale video generative models},
  author={Wan, Team and Wang, Ang and Ai, Baole and Wen, Bin and Mao, Chaojie and Xie, Chen-Wei and Chen, Di and Yu, Feiwu and Zhao, Haiming and Yang, Jianxiao and others},
  journal={arXiv preprint arXiv:2503.20314},
  year={2025}
}

@misc{klingteam2025klingomnitechnicalreport,
      title={Kling-Omni Technical Report}, 
      author={Kling Team and Jialu Chen and Yuanzheng Ci and Xiangyu Du and Zipeng Feng and Kun Gai and Sainan Guo and Feng Han and Jingbin He and Kang He and Xiao Hu and Xiaohua Hu and Boyuan Jiang and Fangyuan Kong and Hang Li and Jie Li and Qingyu Li and Shen Li and Xiaohan Li and Yan Li and Jiajun Liang and Borui Liao and Yiqiao Liao and Weihong Lin and Quande Liu and Xiaokun Liu and Yilun Liu and Yuliang Liu and Shun Lu and Hangyu Mao and Yunyao Mao and Haodong Ouyang and Wenyu Qin and Wanqi Shi and Xiaoyu Shi and Lianghao Su and Haozhi Sun and Peiqin Sun and Pengfei Wan and Chao Wang and Chenyu Wang and Meng Wang and Qiulin Wang and Runqi Wang and Xintao Wang and Xuebo Wang and Zekun Wang and Min Wei and Tiancheng Wen and Guohao Wu and Xiaoshi Wu and Zhenhua Wu and Da Xie and Yingtong Xiong and Yulong Xu and Sile Yang and Zikang Yang and Weicai Ye and Ziyang Yuan and Shenglong Zhang and Shuaiyu Zhang and Yuanxing Zhang and Yufan Zhang and Wenzheng Zhao and Ruiliang Zhou and Yan Zhou and Guosheng Zhu and Yongjie Zhu},
      year={2025},
      eprint={2512.16776},
      archivePrefix={arXiv},
      primaryClass={cs.CV},
      url={https://arxiv.org/abs/2512.16776}, 
}
\bibliographystyle{icml2025}

\newpage
\appendix
\onecolumn
\section{Additional Ablation Studies}
\label{app:abla details}
\subsection{Ablation on Geometric Isolation Phase ($\tau_g$).}
To validate the necessity of the Geometric Isolation Phase described in \cref{sec:3.2}, we conduct an ablation study on the timing threshold $\tau_g$. This parameter controls how long the illumination fusion weight $\lambda(t)$ is forced to zero, allowing the EX-4D backbone to establish structure without interference from the lighting prior.

\begin{figure*}[t]
\vskip 0.2in
\begin{center}
\includegraphics[width=\textwidth]{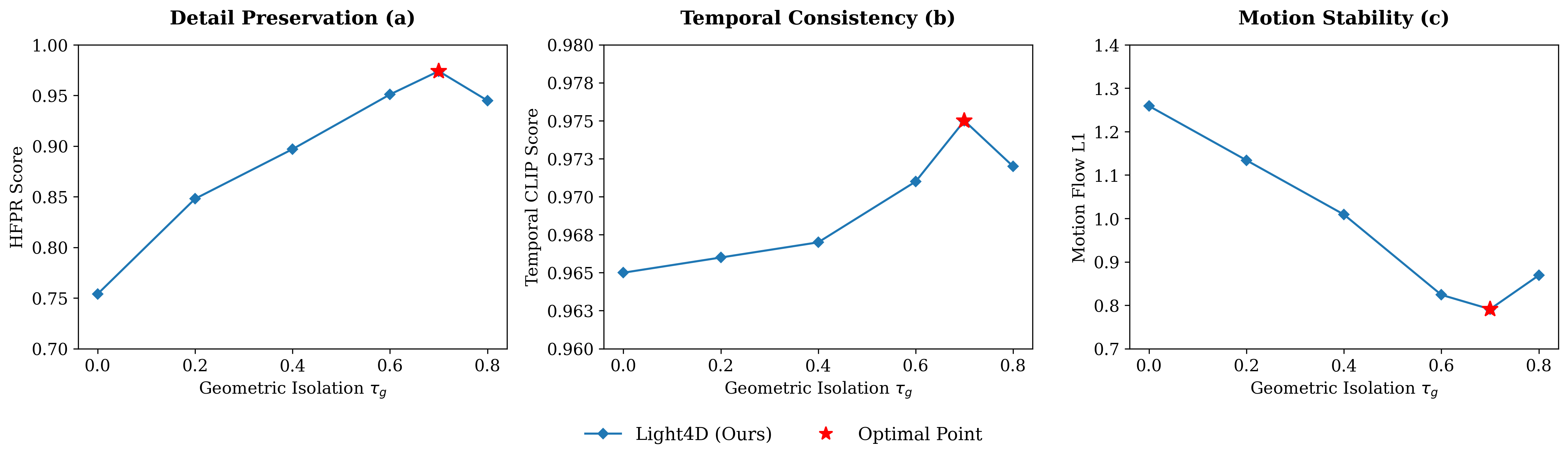}
\caption{\textbf{Quantitative ablation on the geometric isolation threshold $\tau_g$.} (a) HFPR scores, (b) Temporal CLIP scores, and (c) Motion Flow L1 errors are shown.}
\label{fig:ablation_graph}
\end{center}
\vskip -0.2in
\end{figure*}

A quantitative comparison of geometric isolation thresholds $\tau_g$ is provided in \cref{fig:ablation_graph}. We use all evaluation metrics, including video quality and relighting metrics. The results indicate a clear trade-off: premature lighting injection can disrupt the formation of coherent 4D geometry, often leading to reduced temporal consistency and visible flickering, as evidenced by higher motion errors and lower detail preservation. Conversely, excessively delayed injection reduces the number of denoising steps available for illumination integration, potentially resulting in incomplete or unnatural lighting effects. Overall, our results suggest that setting $\tau_g = 0.7$ offers a favorable balance, anchoring the 4D structure while leaving sufficient flexibility for the relighting prior.

\section{Deterministic Coherence and Regularization}
\label{app:det_coh}
To eliminate temporal artifacts, we apply deterministic regularization to the raw predictions $\hat{x}_0^{light}$ from the IC-Light prior before integration. Instead of directly injecting the stochastic output, we first rectify the signal to improve temporal stability. This processed appearance is then fused with the geometric projection to construct the hybrid flow target.

\textbf{Canonical Noise Initialization.} The stochastic nature of diffusion sampling introduces aleatoric uncertainty, leading to high-frequency texture jitter across views. We mitigate this by conditioning the relighting model on a canonical noise prior. Instead of independent sampling, we broadcast a fixed noise map $\epsilon_{\text{shared}}$ across the temporal sequence, ensuring a topologically consistent generation path for identical semantic regions.
\begin{equation}
\epsilon_f = \epsilon_{\text{shared}}, \quad \forall f \in [1, F]
\label{eq:cni}
\end{equation}
where $F$ denotes the number of frames in the video.

\textbf{Global Moment Matching.} To address global exposure fluctuations caused by frame-independent processing, we enforce statistical alignment across the video sequence. We construct a canonical reference by computing the temporal average of the entire prediction sequence. Specifically, we rectify the intensity distribution of each raw prediction $\hat{x}^{light}_{f}$ to match the mean $\mu_{ref}$ and standard deviation $\sigma_{ref}$ of this temporal average reference. This ensures consistent brightness levels throughout the trajectory, yielding the aligned intermediate frame $\tilde{x}_f$:
\begin{equation}
\tilde{x}_f = \mu_{ref} + \frac{\sigma_{ref}}{\sigma_f} (\hat{x}^{light}_{f} - \mu_f)
\label{eq:gmm}
\end{equation}
where $\mu_f$ and $\sigma_f$ denote the intensity mean and standard deviation of the current frame $f$, respectively.

\textbf{Frequency-Decoupled Illuminance Regularization.} Finally, to resolve luminance flickering without compromising geometric sharpness, we employ a spectral decomposition strategy. Based on the assumption that valid illumination changes are spectrally concentrated in the low-frequency domain while texture resides in high frequencies, we decompose the aligned signal $\tilde{x}_f$. We apply a temporal smoothing operator $\mathcal{T}$ exclusively to the base illumination layer, while preserving the high-frequency details:
\begin{equation}
x^{light}_f = \underbrace{\mathcal{T}(\tilde{x}_f * G_\sigma)}_{\text{Smoothed Illumination}} + \underbrace{(\tilde{x}_f - \tilde{x}_f * G_\sigma)}_{\text{Preserved Texture}}
\label{eq:fdi}
\end{equation}
where $*$ denotes the convolution operation. The resulting $x^{light}_f$ serves as the final robust appearance target for the flow matching objective defined in Eq. (2).

\begin{table*}[ht]
\caption{\textbf{Ablation Study:} Summary of relighting and video quality metrics at a $30^{\circ}$ viewpoint change. Best results are \textbf{bolded}, and second-best results are \underline{underlined}.}
\label{tab:ablation_relighting_30_app}
\vskip 0.15in
\begin{center}
\begin{small}
\begin{sc}
\renewcommand{\arraystretch}{0.95}
\begin{tabular*}{\textwidth}{l @{\extracolsep{\fill}} c c c c c c c}
\toprule
Method & CLIP $\uparrow$ & Motion $\downarrow$ & HFPR $\uparrow$ & Aes $\uparrow$ & PSNR $\uparrow$ & SSIM $\uparrow$ & LPIPS $\downarrow$ \\
\midrule
w/o CLA & 0.972& 0.903 & 0.882& 0.229 & \underline{13.698}& \underline{0.739} & \underline{0.420} \\
w/o DGA & 0.965& 1.259 & 0.754 & 0.213 & 10.767 & 0.513 & 0.718 \\
\midrule
w/o FDI & 0.970& 0.885& 0.910& 0.228 & 13.351& 0.680 & 0.502\\
w/o CNI & 0.971& 0.875& 0.925& 0.229 & 13.487& 0.783& 0.470 \\
w/o GMM & 0.972& 0.860& 0.940& 0.230 & 13.650 & 0.711& 0.461\\
w/o All & \underline{0.793}& \underline{0.898} & \underline{0.893}& \underline{0.231} & 13.207 & 0.657 & 0.518 \\
\midrule
\textbf{Full Model} & \textbf{0.975}& \textbf{0.791}& \textbf{0.974}& \textbf{0.243} & \textbf{14.056} & \textbf{0.761} & \textbf{0.360}\\
\bottomrule
\end{tabular*}
\end{sc}
\end{small}
\end{center}
\vskip -0.1in
\end{table*}

\section{More Implementation Details}
\label{app:impl_details}

\subsection{Baselines.}
We evaluate \textbf{\textit{Light4D}} against both training-based and training-free baselines under identical extreme viewpoint changes ($30^{\circ}$, $90^{\circ}$, and $180^{\circ}$). To ensure a fair comparison, we align the camera trajectories used for video generation across methods. Specifically, for Light-X~\cite{liu2025light}, we adopt its original camera extrinsic sequence with an EX-4D-style trajectory parameterization that matches both the viewpoint range and the motion pattern used in our setup, so that all methods are evaluated under the same camera motion. For Light-A-Video (LAV)~\cite{zhou2025light}, we use the officially released implementation of Wan2.1~\cite{wan2025wan}.

\subsection{Time-aware fusion schedule.}
We use a piecewise schedule for the fusion weight: it is set to $0$ for the first $60\%$ of denoising steps (Steps $0$--$14$) to prioritize geometric completion, and then linearly increases to a maximum value of $0.5$ during the final $40\%$ (Steps $15$--$24$) for progressive illumination injection.

\subsection{Temporal consistency and smoothing.}
The Consistent Light Attention (CLA) module uses a balance factor of $\gamma=0.7$. The final output is processed with adaptive temporal smoothing using a window size of $9$ and a Gaussian kernel with $\sigma=25$ to suppress high-frequency flicker while preserving texture details.

\subsection{User Study and Preference}
\begin{figure}
    \centering
    \includegraphics[width=0.85\linewidth]{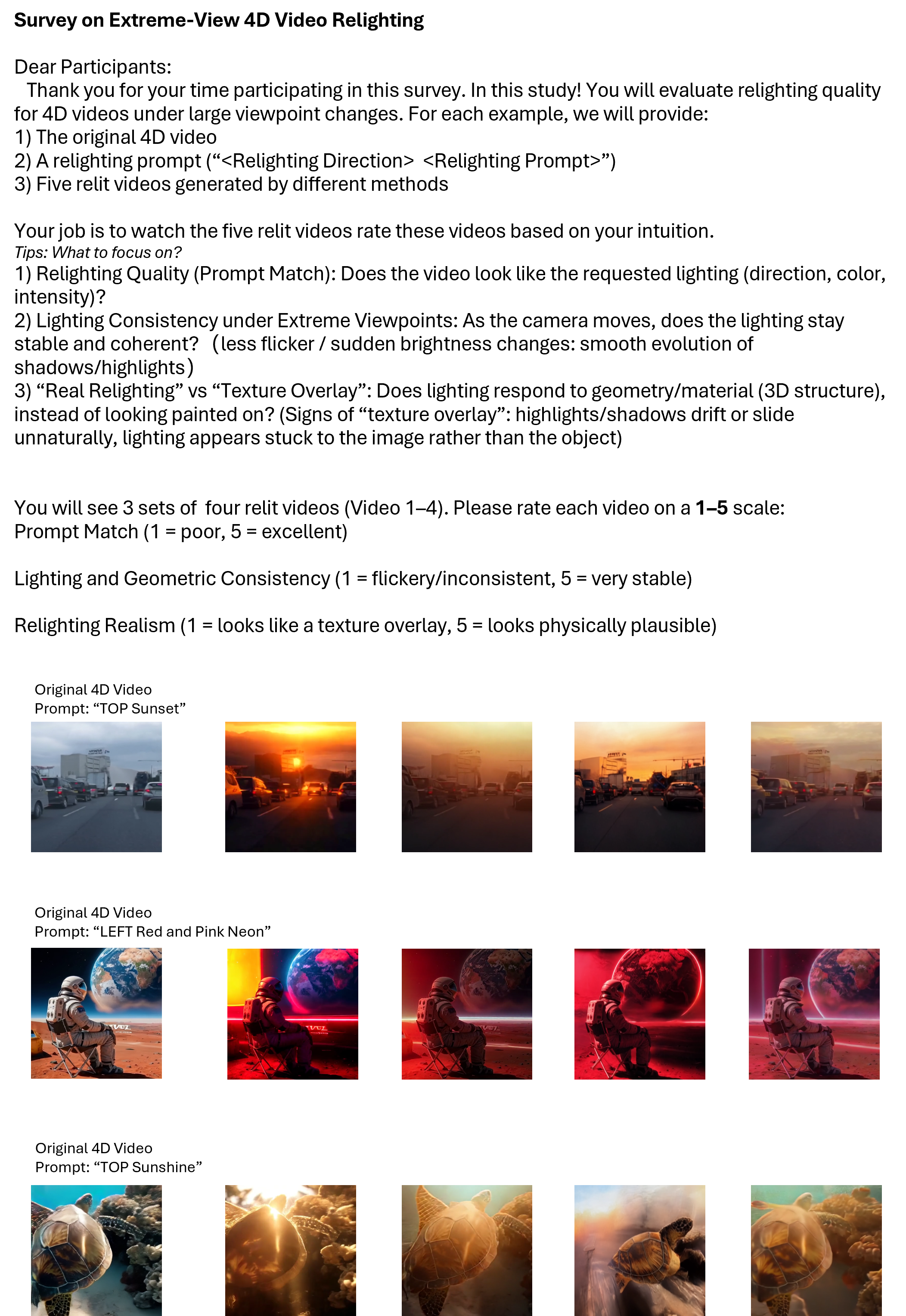}
    \caption{Example of User Study Survey}
    \label{fig:userstudysurvey}
\end{figure}

We select three test sequences with viewpoint-change ranges of $30^{\circ}$, $90^{\circ}$, and $180^{\circ}$.
For each viewpoint setting, we generate relit videos using three baselines~\cite{hu2025ex,zhou2025light,zhang2025scaling,liu2025light} and our method, all conditioned on the same source 4D video and the same relighting prompt.
We distribute the survey to 30 participants and collect their ratings for each relit video on a 1--5 Likert scale along four dimensions:
(i) Prompt Match (1 = poor, 5 = excellent),
(ii) Lighting Consistency (1 = flickery, inconsistent, 5 = very stable),
(iii) Geometric Consistency (1 = unstable, 5 = very stable),
and (iv) Relighting Realism (1 = fake or overlay-like, 5 = realistic lighting).
Figure~\ref{fig:userstudysurvey} shows an example survey interface used in our study, where participants are provided with the source 4D video, the relighting prompt, and the five anonymized relighting results for side-by-side comparison.
Results are presented in Table~\ref{tab:user_study}.

\begin{table*}[t]
\caption{\textbf{User Study.} Ratings are on a 1 to 5 Likert scale.
PM: Prompt Match; LC: Lighting Consistency; GC: Geometric Consistency; RR: Relighting Realism.
Best results are \textbf{bolded}; second best are \underline{underlined}.}
\label{tab:user_study}
\vskip 0.15in
\centering
\renewcommand{\arraystretch}{1.05}
\setlength{\tabcolsep}{4.2pt}
\begin{small}
\begin{tabular}{l c c c c c c c c c c c c}
\toprule
& \multicolumn{4}{c}{$30^{\circ}$} & \multicolumn{4}{c}{$90^{\circ}$} & \multicolumn{4}{c}{$180^{\circ}$} \\
\cmidrule(lr){2-5}\cmidrule(lr){6-9}\cmidrule(lr){10-13}
Method
& PM$\uparrow$ & LC$\uparrow$ & GC$\uparrow$ & RR$\uparrow$
& PM$\uparrow$ & LC$\uparrow$ & GC$\uparrow$ & RR$\uparrow$
& PM$\uparrow$ & LC$\uparrow$ & GC$\uparrow$ & RR$\uparrow$ \\
\midrule
\multicolumn{13}{l}{\textit{Training-based}}\\
Light-X
& \underline{4.2} & \underline{4.0} & 4.1 & 1.6
& \underline{4.0} & \underline{3.7} & 3.9 & 1.4
& \underline{3.6} & \underline{2.9} & 3.5 & 1.2 \\
\midrule
\multicolumn{13}{l}{\textit{Training-free}}\\
EX-4D + IC-Light
& 4.0 & 1.8 & \underline{4.2} & 2.2
& 3.8 & 1.5 & \underline{4.0} & 2.0
& 3.5 & 1.3 & \underline{3.7} & 1.8 \\
EX-4D + LAV
& 3.8 & 3.6 & 3.2 & \underline{3.1}
& 3.6 & 3.3 & 2.9 & \underline{2.8}
& 3.3 & 2.8 & 2.5 & \underline{2.4} \\
Light4D (Ours)
& \textbf{4.5} & \textbf{4.4} & \textbf{4.3} & \textbf{4.4}
& \textbf{4.4} & \textbf{4.2} & \textbf{4.2} & \textbf{4.2}
& \textbf{4.2} & \textbf{3.9} & \textbf{4.0} & \textbf{3.9} \\
\bottomrule
\end{tabular}
\end{small}
\vskip -0.1in
\end{table*}

\section{Limitation and Future Work}
As a training-free framework, Light4D's performance is bounded by the capabilities of its foundation models---specifically, the geometric fidelity of EX-4D and the single-frame nature of the IC-Light prior. Consequently, maintaining rigorous global lighting consistency remains challenging when adapting 2D image priors to the temporal domain, particularly during extreme viewpoint traversals ($-90^{\circ}\sim90^{\circ}$). Future work will leverage the framework's modularity to integrate more advanced video-native illumination models and stronger 4D generative backbones to further improve synthesis quality. We also aim to investigate specialized mechanisms to mitigate photometric inconsistencies induced by extreme viewpoint changes.

\section{More Experimental Results}

\begin{algorithm*}[tb]
   \caption{Inference of Light4D}
   \label{alg:inference}
\begin{algorithmic}[1]
\STATE {\bfseries Input:} source video $\mathcal{V}_s$, target camera $\mathcal{C}$, lighting prompts $\mathcal{L}$, EX-4D model $\mathbf{V}_{geo}$, IC-Light prior $\mathcal{M}_{light}$, VAE $(\mathcal{E}, \mathcal{D})$, noise levels $\{\sigma_k\}$, time schedule $\{t_k\}$
    \STATE {\bfseries Output:} relit 4D video $\mathcal{V}$
    \STATE Sample noise $\epsilon_{shared}$ \hfill // Canonical Noise Initialization (\cref{eq:cni})
    \STATE $z \leftarrow z_{init}(\mathcal{V}_s, \mathcal{C})$ \hfill // Initialize latent
    \FOR{$k=1, 2, \dots, K$}
    \STATE $t \leftarrow t_k$ \hfill // Map discrete step $k$ to continuous time
    \STATE $\hat{z}_0^{geo} \leftarrow \mathbf{V}_{geo}(z, \sigma_t )$ \hfill // Estimate geometry state
    \IF{$\lambda(t) > 0$} 
    \STATE $x^{geo} \leftarrow \mathcal{D}(\hat{z}_0^{geo})$ \hfill // Decode to RGB space
    \STATE $\hat{x}_0^{light} \leftarrow \mathcal{M}_{light}(x^{geo}, \mathcal{L}, \epsilon_{shared}; \text{TCA})$ \hfill // Inject light (\cref{eq:x0_light})
    \STATE $\tilde{x} \leftarrow \text{GMM}(\hat{x}_0^{light})$ \hfill // Global Moment Matching (\cref{eq:gmm})
    \STATE $x^{light} \leftarrow \text{FDI}(\tilde{x})$ \hfill // Freq-Decoupled Regularization (\cref{eq:fdi})
    \STATE $x^{fuse} \leftarrow (1-\lambda(t)) x^{geo} + \lambda(t) x^{light}$
    \STATE $z_{target} \leftarrow \mathcal{E}(x^{fuse})$ \hfill // Hybrid target (\cref{eq:z_target})
    \ELSE
    \STATE $z_{target} \leftarrow \hat{z}_0^{geo}$ \hfill // Geometric Isolation Phase
    \ENDIF
    \STATE $z \leftarrow z + (\sigma_{t'} - \sigma_t) \cdot \frac{z - z_{target}}{\sigma_t + \delta}$ \hfill // Euler solver (\cref{eq:euler_update})
    \ENDFOR
    \STATE $\mathcal{V} \leftarrow \mathcal{D}(z)$ 
    \STATE \bfseries{return} $\mathcal{V}$
\end{algorithmic}
\end{algorithm*}

In this appendix, we provide additional qualitative results that highlight the effectiveness of our method under challenging viewpoints, diverse lighting prompts, and complex real-world environments.
\subsection{Visualizations under Normal Viewpoints.}
\begin{figure*}[t]
\vskip 0.2in
\begin{center}
\includegraphics[width=\textwidth]{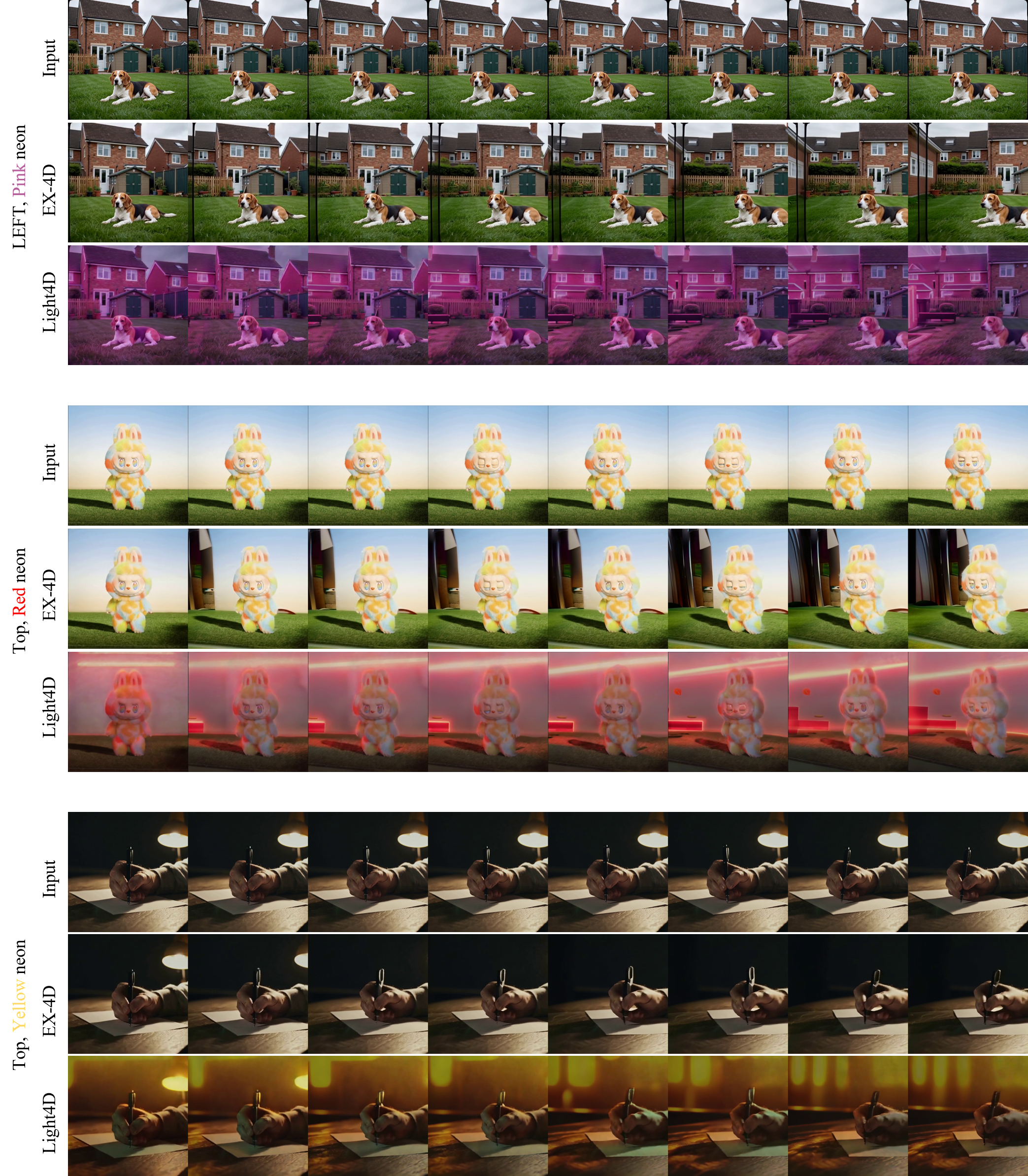}
\caption{\textbf{Extended Qualitative Results Under Normal Viewpoints.} Our method generates realistic light-material interactions and smooth temporal transitions over diverse object categories.}
\label{fig:ablation_app}
\end{center}
\vskip -0.2in
\end{figure*}

\begin{figure*}[t]
\vskip 0.2in
\begin{center}
\includegraphics[width=\textwidth]{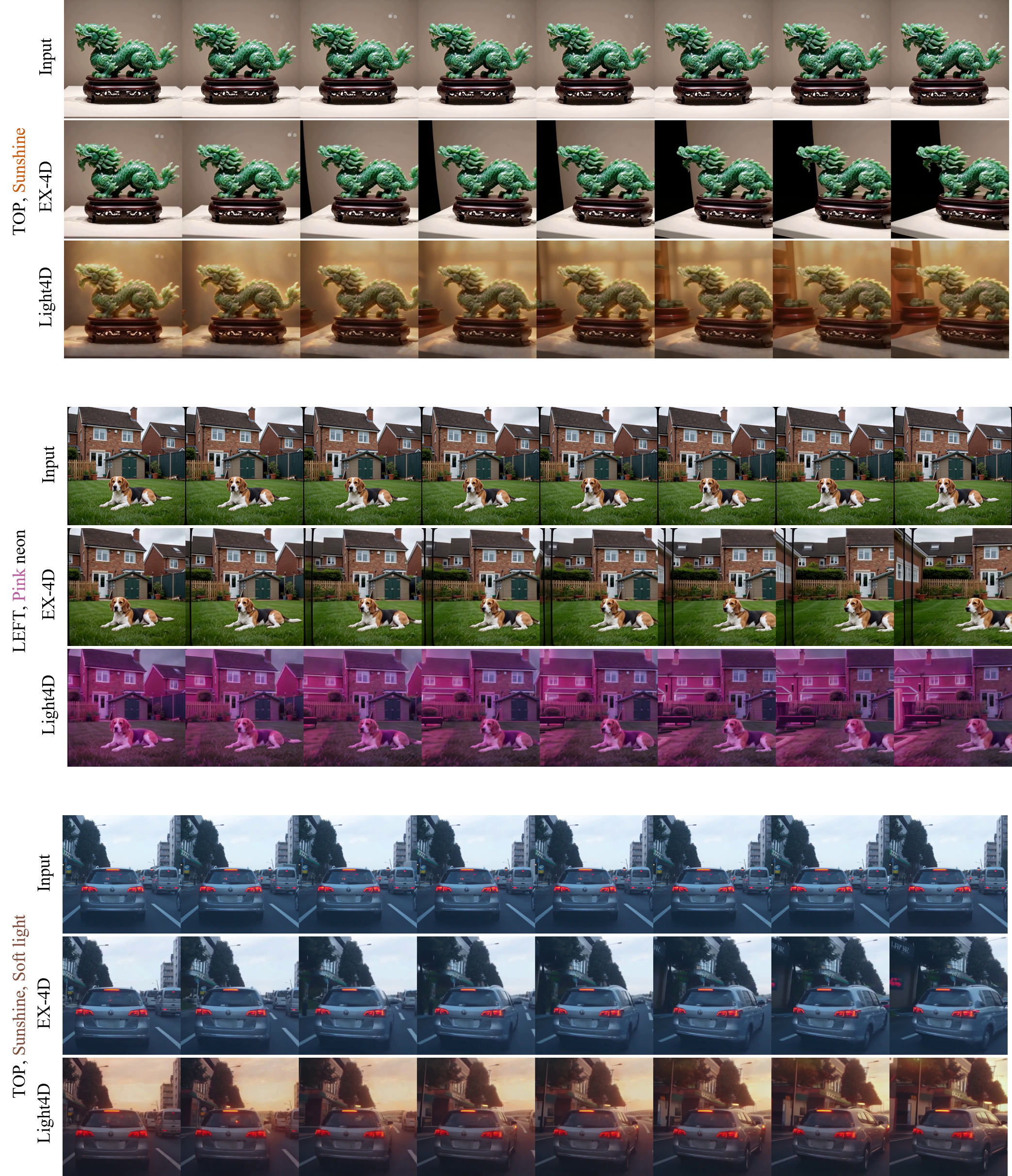}
\caption{
\textbf{Another Extended Qualitative Results Under Normal Viewpoints.} Our method generates realistic light-material interactions and smooth temporal transitions over diverse object categories.}
\label{fig:ablation_30demo}
\end{center}
\vskip -0.2in
\end{figure*}

We show qualitative comparisons across a variety of object categories under standard viewing conditions.
As shown in Figures~\ref{fig:ablation_app} and~\ref{fig:ablation_30demo}, these results highlight high-frequency detail preservation and natural light--material interactions.
Our method also produces smooth temporal transitions as the light source moves.
With our deterministic coherence and regularization mechanisms, we suppress the temporal flicker commonly observed in diffusion-based baselines, yielding stable 4D sequences.

\subsection{Visualizations under Extreme Viewpoints.}
\begin{figure*}[t]
\vskip 0.2in
\begin{center}
\includegraphics[width=\textwidth]{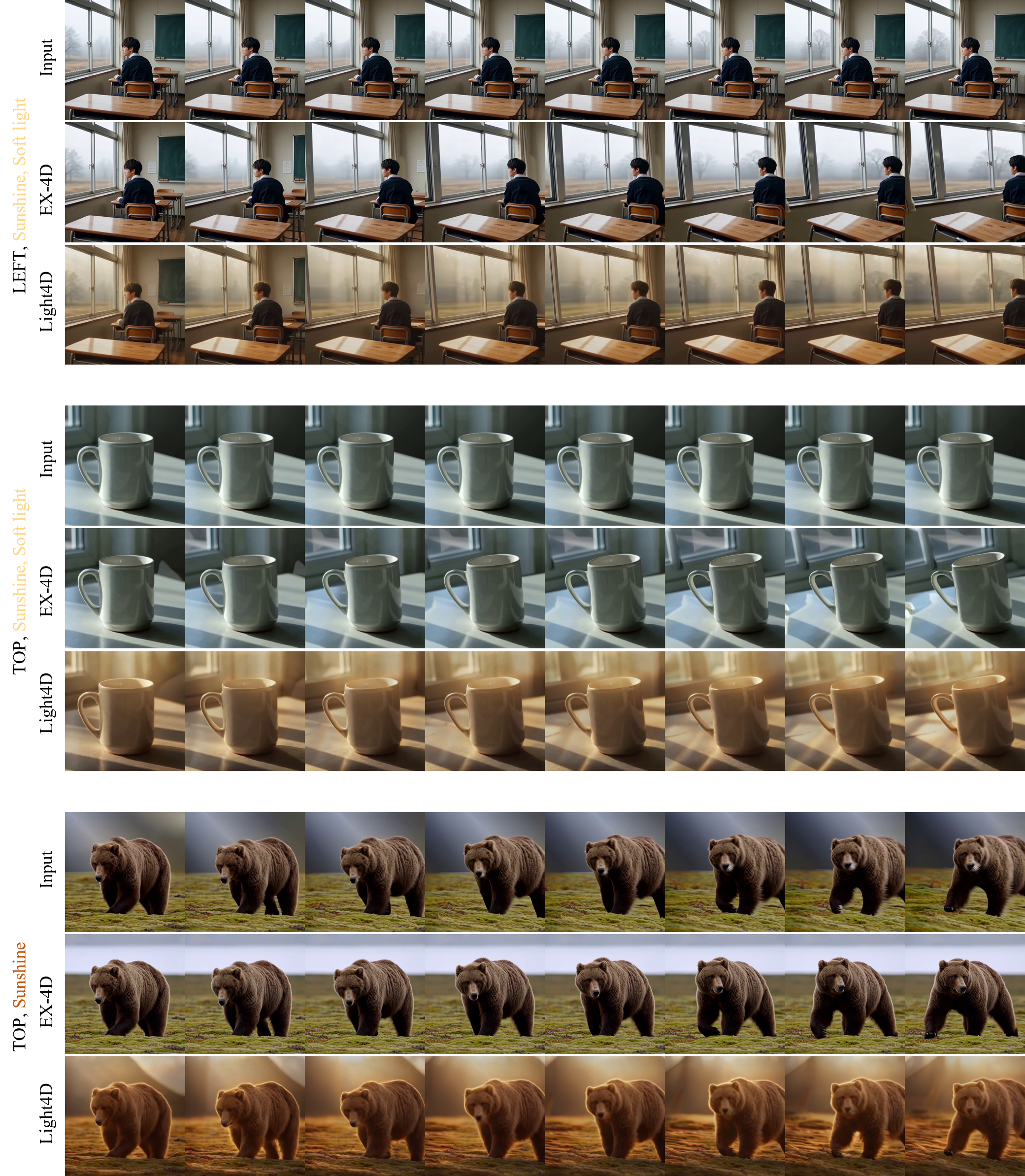}
\caption{\textbf{Extended qualitative results under extreme viewpoints.} Our method shows strong geometric robustness and maintains stable 3D structure even under large camera viewpoint shifts.}
\label{fig:ablation_180demo}
\end{center}
\vskip -0.2in
\end{figure*}

To evaluate geometric robustness, we visualize relighting results under large camera viewpoint shifts.
These settings are challenging because the overlap between reference and target views is small, which can cause geometric distortion or texture drift in prior methods.
As shown in Figure~\ref{fig:ablation_180demo}, our method preserves the underlying 3D structure without noticeable ``ghosting'' artifacts.
By decoupling the geometric isolation phase from illumination modulation, our approach keeps relit shadows and highlights spatially coherent even under large viewpoint changes.

\subsection{Visualizations in Autonomous Driving Scenarios.}
\begin{figure*}[t]
\vskip 0.2in
\begin{center}
\includegraphics[width=\textwidth]{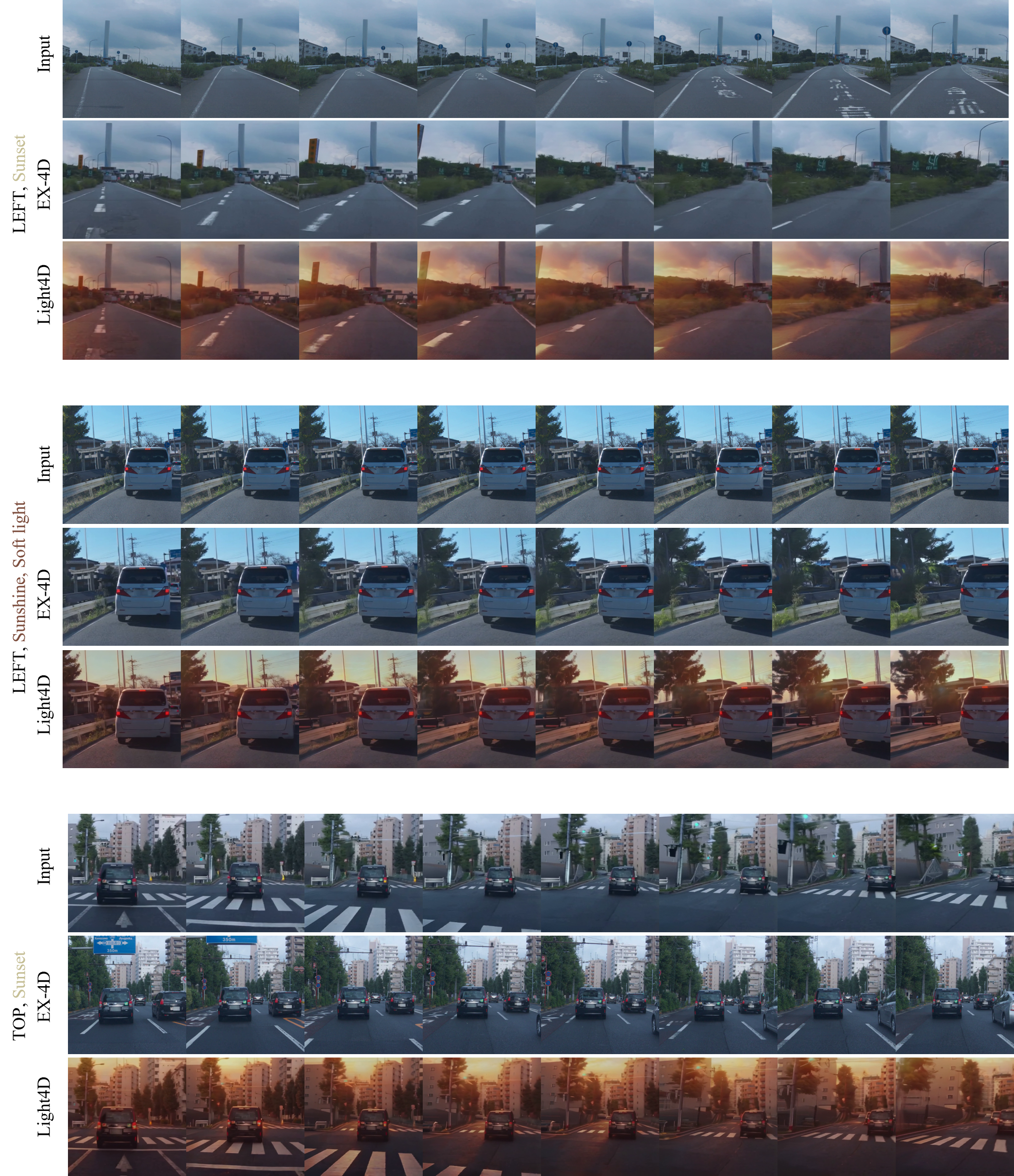}
\caption{\textbf{Extended qualitative results in autonomous driving scenarios.} Our method generalizes well to complex outdoor environments and produces realistic shadows for dynamic vehicles.}
\label{fig:ablation_AD}
\end{center}
\vskip -0.2in
\end{figure*}

To demonstrate practical utility on real videos, we apply \textbf{\textit{Light4D}} to real-world autonomous driving sequences from OpenScene~\cite{OpenScene_Dataset_Contributors_OpenScene_The_Largest_2023}.
These scenes include dynamic foreground objects and large outdoor environments.
As shown in Figure~\ref{fig:ablation_AD}, our method handles outdoor lighting changes and produces realistic shadows on moving vehicles, highlighting its potential for high-fidelity relighting and data augmentation in driving scenarios.

\end{document}